\documentclass[lettersize,journal]{IEEEtran}
\usepackage{amsmath,amsfonts, amssymb}
\usepackage{algorithmic}
\usepackage{algorithm}
\usepackage{array}
\usepackage{textcomp}
\usepackage{stfloats}
\usepackage{url}
\usepackage{verbatim}
\usepackage{graphicx, subfig}
\usepackage{cite}
\usepackage{color}
\usepackage{bm}
\begin{document}

\title{Practical Physical Layer Authentication for Mobile Scenarios Using a Synthetic Dataset Enhanced Deep Learning Approach}

\author{Yijia Guo,~\IEEEmembership{Graduate Student Member,~IEEE,}
        Junqing~Zhang,~\IEEEmembership{Senior Member,~IEEE},\\
        and Y.-W. Peter Hong,~\IEEEmembership{Senior Member,~IEEE}
        % <-this % stops a space
\thanks{Manuscript received xxx; revised xxx; accepted xxx. Date of publication xxx; date of current version xxx. The work of J. Zhang was supported in part by the UK EPSRC under grant ID EP/V027697/1 and EP/Y037197/1, and in part by Royal Society Research Grants under grant ID RGS/R1/231435. The work of Y.-W.~P.~Hong was supported in part by the National Science and Technology Council (NSTC) of Taiwan under grant NSTC 111-2221-E-007-042-MY3. The review of this paper was coordinated by xxx. 
    \textit{(Corresponding author: Junqing Zhang.)}}
    \thanks{Y.~Guo is with the Department of Electrical Engineering and Electronics, University of Liverpool, Liverpool, L69 3GJ, United Kingdom. She is also with the Institute of Communications Engineering, National Tsing Hua University, Hsinchu, Taiwan 300044. (email: yijia.guo@liverpool.ac.uk)}
	\thanks{J.~Zhang is with the Department of Electrical Engineering and Electronics, University of Liverpool, Liverpool, L69 3GJ, United Kingdom. (email: junqing.zhang@liverpool.ac.uk)}
    \thanks{Y.-W. P. Hong is with the Institute of Communications Engineering, National Tsing Hua University, Hsinchu, Taiwan 300044. (email:ywhong@ee.nthu.edu.tw)}    
	\thanks{Color versions of one or more of the figures in this paper are available online at http://ieeexplore.ieee.org.}
	\thanks{Digital Object Identifier xxx}	
}

\markboth{Journal of \LaTeX\ Class Files,~Vol.~14, No.~8, August~2021}%
{Shell \MakeLowercase{\textit{et al.}}: A Sample Article Using IEEEtran.cls for IEEE Journals}


\maketitle

\begin{abstract}
The Internet of Things (IoT) is ubiquitous thanks to the rapid development of wireless technologies. However, the broadcast nature of wireless transmissions results in great vulnerability to device authentication. Physical layer authentication emerges as a promising approach by exploiting the unique channel characteristics. However, a practical scheme applicable to dynamic channel variations is still missing. In this paper, we proposed a deep learning-based physical layer channel state information (CSI) authentication for mobile scenarios and carried out comprehensive simulation and experimental evaluation using IEEE 802.11n. Specifically, a synthetic training dataset was generated based on the WLAN TGn channel model and the autocorrelation and the distance correlation of the channel, which can significantly reduce the overhead of manually collecting experimental datasets.  A convolutional neural network (CNN)-based Siamese network was exploited to learn the temporal and spatial correlation between the CSI pair and output a score to measure their similarity. We adopted a synergistic methodology involving both simulation and experimental evaluation. The experimental testbed consisted of WiFi IoT development kits and a few typical scenarios were specifically considered. Both simulation and experimental evaluation demonstrated excellent generalization performance of our proposed deep learning-based approach and excellent authentication performance. Demonstrated by our practical measurement results, our proposed scheme improved the area under the curve (AUC) by 0.03 compared to the fully connected network-based (FCN-based) Siamese model and by 0.06 compared to the correlation-based benchmark algorithm.
 % our deep learning-based scheme always outperforms the correlation-based benchmark algorithm, with an average gain of 0.06 in the area under curve (AUC).
\end{abstract}


\section{Introduction}
\label{sec:introduction}
\IEEEPARstart{T}{he} rapid development of wireless technologies has enabled ubiquitous Internet of Things (IoT) connectivity and triggered numerous transformative applications to our everyday life, e.g., smart home, smart cities, connected healthcare, industrial IoT~\cite{al2015internet}.
Such a revolution is enabled by massively connected devices through wireless communication technology, whose number is predicted by International Data Corporation (IDC) to reach 55.7 billion by 2025\footnote{https://blogs.idc.com/2021/01/06/future-of-industry-ecosystems-shared-data-and-insights/}. Wireless communications are preferred to connect them, such as WiFi, Bluetooth, ZigBee, and LoRa.
However, due to the broadcast nature of the wireless medium, any device within the communication range can get access to the signal. This results in spoof attacks being carried out easily, where an attacker pretends to be a legitimate user. 
Device authentication serves as the countermeasure, which identifies device identity to allow network access for legitimate devices and deny malicious users. Existing authentication schemes rely on media access control (MAC) address as the identifier, which can however be easily spoofed. 

There has been growing interest in physical layer authentication which identifies a device using its channel characteristics~\cite{wang2020physical,xie2020survey,hoang2024physical}. It can be categorized into received signal strength (RSS)-based and channel state information (CSI)-based schemes. RSS is utilized to detect the existence of rogue devices~\cite{yang2009detecting}, determine the number of attackers~\cite{yang2013detection}, and locate the adversaries~\cite{chen2010detecting}. However, RSS is a coarse-grained estimate of the channel, hence the detection accuracy is limited. In contrast, CSI is finer-grained, which can provide more detailed information of the channel. Therefore, CSI, including channel frequency response (CFR)~\cite{xiao2007fingerprints, xiao2008mimo-assisted, xiao2008using, xiao2009channel} and channel impulse response (CIR)~\cite{Tugnait2010channel-based, liu2011robust, liu2013two, liu2016physical, xie2021physical, fang2020fuzzy, zhang2021exploiting}, has been widely used to enhance the authentication performance.

Most of the existing work focuses on stationary scenarios, where all the devices remain at their fixed places and no other channel variation is caused by the environment. In this case, a receiver will always obtain constant channel characteristics from devices. A CFR-based scheme is designed in~\cite{xiao2007fingerprints}, which is extended by multiple-input and multiple-output (MIMO) to get a security gain in~\cite{xiao2008mimo-assisted}. A CIR-based scheme is proposed for single-carrier wireless networks in~\cite{Tugnait2010channel-based}. Since the channel can be used as a unique fingerprint of devices, many learning-based methods have been proposed to classify the channel patterns and identify device identity. $K$-means algorithm is applied in~\cite{liu2014practical}, where the cluster number of CSI is utilized to determine whether there is an attacker. Support vector machine (SVM) is used to obtain the similarity between the unknown CSI and the local user profile for device authentication~\cite{shi2017smart}. 
Deep learning techniques, such as convolutional neural network (CNN), are also used, e.g., in~\cite{liao2019novel}. 

However, the dynamic channel is not considered in the above work, which significantly limits their applications in practical IoT scenarios. Many devices, e.g. smartphones, will be mobile. In other cases, IoT devices, e.g., smart meters, may remain fixed, but the surrounding wireless environment will vary due to passing pedestrians and/or vehicles. When the channel is dynamic due to device moving and/or environment variations, the channel characteristics measured at each device are time-varying. Unfortunately, the varied channel characteristics cannot be considered as a unique pattern anymore, which invalidates the approaches for stationary scenarios.

Early research on physical layer authentication in mobile scenarios was primarily based on hypothesis testing.
A CFR-based scheme is proposed in~\cite{xiao2008using} considering the channel correlations among the time, frequency and spatial domains. Moreover, its application in frequency-selective Rayleigh channels is studied in~\cite{xiao2009channel}. The CIR-based scheme is proposed for wireless communications in a time-varying multipath channel in~\cite{liu2011robust}. Then the CIR-based scheme is integrated with multipath delay for reliable authentication performance at low signal-to-noise ratio (SNR) conditions~\cite{liu2013two}. Followed by~\cite{liu2016physical}, the CIR-based scheme is further combined with a two-dimensional quantization method to simplify the decision rule for authentication. In recent years, several multi-dimensional authentication mechanisms have been proposed. A multiple CIRs physical layer authentication scheme is proposed to reduce the performance loss due to quantization error~\cite{xie2021physical}. Location-specific channel gain and transmitter-specific phase noise are exploited for physical layer authentication in massive MIMO systems~\cite{zhang2021exploiting}. The above works mainly provide hypothesis testing-based modeling and analysis. In addition to the hypothesis testing-based studies, the temporal correlation in the channel is leveraged in~\cite{liu2018authenticating}, where Pearson correlation is computed and authentication is performed based on an empirically determined threshold.

With the development of data-driven technologies, many learning-based methods have been proposed, which can be categorized into three approaches. 
The first approach involves authentication based on channel prediction. In~\cite{germain2021channel}, Long Short-Term Memory (LSTM) and Gated Recurrent Units (GRU) are investigated for MIMO systems to predict future channel states using previous channel measurements, followed by threshold-based mean square error (MSE) detection. Similarly, in~\cite{wang2021channel}, the legitimate CSI is predicted based on historical CSI and the transmitter's geographical information for authentication. The work assumes the trajectory of the device is either coordinated or even controlled, which is unrealistic in practical scenarios. Moreover, these studies lack experimental validation with real-world systems.
The second approach focuses on authentication using classification algorithms, where legitimate and rogue devices move in distinct regions, ensuring non-overlapping CSI measurements to establish a clear classification boundary. Decision Trees (DT), Support Vector Machines (SVM), K-Nearest Neighbors (KNN), and ensemble learning are explored in~\cite{pan2019threshold} to classify CSI measurements in mobile scenarios. Furthermore, a weighted voting scheme based on the SVM classifier from~\cite{pan2019threshold} is introduced in~\cite{xie2021weighted} to improve classification accuracy. A ResNet model is implemented to extract features from mobile CSI measurements in~\cite{jing2024multi-user}. However, in real-world scenarios, if the paths of legitimate and rogue devices overlap and their channels are similar, classification-based methods are no longer applicable. Another classification-based algorithm was proposed in~\cite{fang2020fuzzy}, where fuzzy learning is used to indicate the likelihood that CSI and other physical layer information belong to each class.
The third approach is similarity-based authentication.  In~\cite{zhang2025enhancing}, a method is proposed that combines a sliding window with a Siamese network, with a fully connected network (FCN) as the embedding network.

In this paper, we proposed a practical and robust deep learning-based physical layer authentication scheme and carried out comprehensive simulation and experimental evaluation. In particular, we carefully tuned a synthetic training dataset to cover the test scenarios, which can significantly eliminate the overhead of collecting experimental datasets. In addition, we adopted a CNN-based Siamese deep learning model to learn the similarity between a pair of CSI estimations. The simulation and experimental evaluation demonstrated the robustness and generalization capability of the proposed scheme. 

To the best of our knowledge, the authentication method proposed in~\cite{liu2018authenticating} and~\cite{zhang2025enhancing} are the most practical ones, hence, they are used as the benchmark algorithms in this paper.
Our main contributions are summarized as follows. 
\begin{itemize}
    \item A synthetic training dataset was generated for indoor mobile scenarios. Specifically, the synthetic channels were created based on the WLAN TGn channel model, along with the autocorrelation and distance correlation of the channel. To ensure the generalization of the training dataset, a comprehensive set of simulation parameters was carefully optimized through simulation evaluation. In the experimental scenarios, the designed synthetic dataset was further used as the training dataset. Validation with experimental test dataset demonstrated that, in various typical indoor scenarios, the synthetic training dataset achieved performance comparable to the experimental training dataset and had the potential to outperform it. This suggested that the offline-generated synthetic training dataset could replace the experimental training dataset for model training, eliminating the need for extensive manual CSI collection. 
    \item A Siamese model utilizing a CNN as the embedding network was employed to capture the temporal and spatial similarity between CSI measurements, and output a score to measure the similarity. The device authentication was achieved by comparing the score with a threshold obtained empirically. It was demonstrated that our scheme could achieve better authentication performance with lower computational overhead compared to the FCN-based Siamese network in~\cite{zhang2025enhancing}.
    % By adopting a deep learning-based approach, our scheme can obtain prior information about the environment and be robust against channel variations.
    \item We carried out a comprehensive simulation evaluation of the proposed scheme on different WLAN channel models in MATLAB. We studied the effect of SNR, the distance between legitimate and rogue devices, and the transmission interval on authentication performance in simulation. The simulation evaluation allowed us to optimize the parameters of the synthetic training dataset quickly and efficiently.
    \item We also performed an extensive experimental evaluation using WiFi in various indoor environments. A testbed using the ESP32 kit and two LoPy4 boards was created. Typical indoor environments involving both line-of-sight (LOS) and non-line-of-sight (NLOS) as well as various SNR were considered.
    We explored the reliability of the synthetic dataset and the effectiveness of the CNN-based Siamese network. The generalization performance of the proposed scheme in different typical test scenarios was also evaluated. Demonstrated by our experimental results, the proposed scheme improved the area under the curve (AUC) by 0.06 compared to the correlation-based benchmark algorithm in~\cite{liu2018authenticating} and by 0.03 compared to the FCN-based Siamese network in~\cite{zhang2025enhancing}.
    % Demonstrated by our experimental results, our deep learning-based scheme always outperformed the correlation-based benchmark algorithm, with an average gain of 0.06 in the area under curve (AUC).
\end{itemize}
The datasets are available online\footnote{https://ieee-dataport.org/documents/wi-fi-channel-state-information-dataset-mobile-physical-layer-authentication}.
In our previous work~\cite{guo2023deep}, we proposed a CNN model for device authentication in mobile scenarios. However, in this work, we 
significantly extend by designing a synthetic training dataset and a Siamese-based model. Moreover, the performance evaluation is conducted and evaluated in both simulation and a real experimental WiFi testbed.

The rest of this paper is organized as follows. Section~\ref{sec:system} introduces the system model and problem statement. Section~\ref{sec:proposed} briefly describes the proposed method for device authentication. Section~\ref{sec:synthetic} elaborates on the generation of the synthetic dataset. Section~\ref{sec:siamese} describes the CNN-based Siamese network. Section~\ref{sec:simulation} and Section~\ref{sec:experiment} present and discuss the simulation results and the experiment results, respectively. Section~\ref{sec:discussion} provides the scalability analysis and advantages of synthetic dataset and Section~\ref{sec:conclusion} concludes the paper.

\section{System Model and Problem Statement}
\label{sec:system}
\subsection{System Model}
\begin{figure}[!t]
\centerline{\includegraphics[width=3.4in]{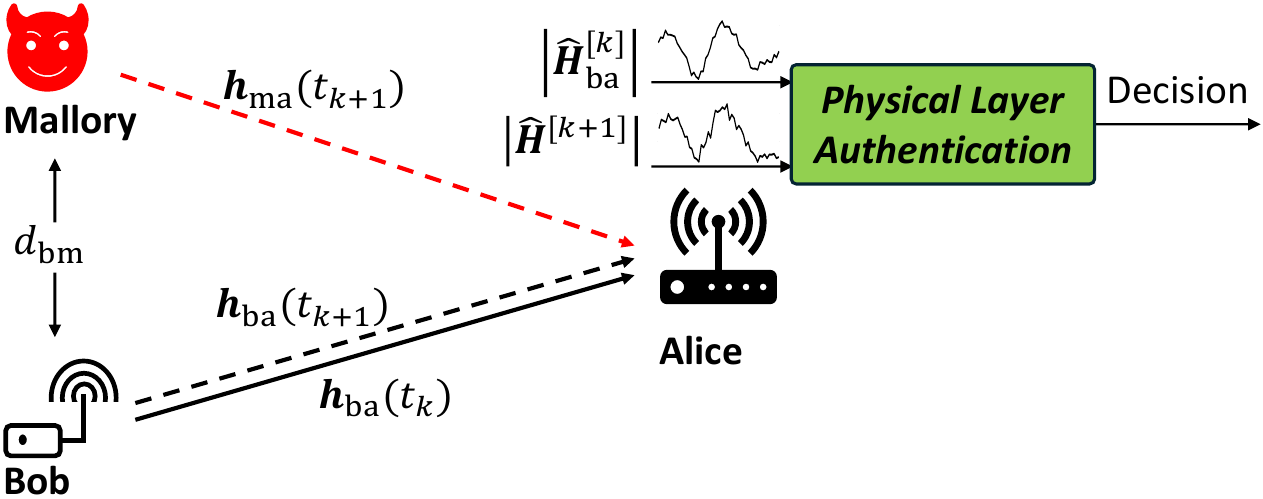}}
\caption{The system model for physical layer authentication. }
\label{fig:SystemModel}
\end{figure}

As shown in Fig.~\ref{fig:SystemModel}, two legitimate users, Alice and Bob,  aim to communicate securely in a time-variant channel.
There is also an attacker, Mallory, who intends to carry out spoof attacks by injecting packets into the open wireless channel. 
All the users are operating in IEEE 802.11n legacy OFDM mode with a $20$ MHz channel bandwidth and $M'=64$ subcarriers. Long training symbols are used for channel estimation which occupy $M=52$ subcarriers. Assuming that Alice receives the $k$-th packet from Bob and $X^{[k]}[m]$ represents the long training symbol modulated to the $m$-th subcarrier in the $k$-th packet, the transmitted time domain signal can be written as
\begin{equation}
x^{[k]}[n]=\sum_{m=0}^{M-1} X^{[k]}[m] e^{j 2 \pi m n / M'},n=0,1,\cdots,N-1.
\end{equation}

% The CIR between Bob and Alice with multipath delay $\tau$ at the transmission time of the $k$-th packet, $t_{k}$, is a dynamic multipath channel and can be given as
% \begin{equation}
% h_{\rm ba}(\tau, t_{k})=\sum_{l=0}^{L-1} h_{\rm ba}(\tau_l, t_{k}) \delta(\tau-\tau_l),
% \end{equation}
% where $L$ denotes the number of the paths, $\tau_l$ represents the delay of the $l$-th path and $\delta(\cdot)$ denotes the Dirac delta function. We denote $h_{\rm ba}^{[k]}[l]=h_{\rm ba}(\tau_l, t_{k})$ as the discrete channel.

The dynamic channel between Bob and Alice at time $t$ has a CIR ${\bm h}_{\rm ba}(t)\triangleq[h_{\rm ba}(0,t),h_{\rm ba}(1,t),\dots,h_{\rm ba}(L-1, t)]$, where $L$ is the number of channel taps. Suppose the $k$-th packet is transmitted at time $t_{k}$, we denote $h_{\rm ba}^{[k]}[l]=h_{\rm ba}(l, t_{k})$ as the corresponding channel.
The $n$-th received signal sample of the long training symbol in the $k$-th packet can be written as
\begin{equation}
y^{[k]}[n] =\sum_{l=0}^{L-1} x^{[k]}[n-l] h_{\rm ba}^{[k]}[l]+z[n],
\end{equation}
where $z[n]$ is the additive Gaussian white noise (AWGN) and $z[n] \sim \mathcal{CN}(0, \sigma_z^2)$. The equivalent frequency domain signal can be written as
\begin{equation}
\begin{aligned}
    Y^{[k]}[m] & =\sum_{n=0}^{M'-1} y^{[k]}[n] e^{-j 2 \pi m n / M'} \\
    & =X^{[k]}[m] H_{\rm ba}^{[k]}[m]+Z[m],
\end{aligned}
\end{equation}
where $H_{\rm ba}^{[k]}[m]$ is the channel coefficient on the $m$-th subcarrier and given as
\begin{equation}
    H_{\rm ba}^{[k]}[m]=\sum_{l=0}^{L-1} h_{\rm ba}^{[k]}[l] e^{-j 2 \pi m l/M'}.
\end{equation}
The estimated channel coefficient over $M$ subcarriers can be written as $\widehat{\bm H}_{\rm ba}^{[k]}=[\widehat{H}_{\rm ba}^{[k]}[0], \widehat{H}_{\rm ba}^{[k]}[1], \dots, \widehat{H}_{\rm ba}^{[k]}[M-1]]$, which we refer to as the CSI and can be obtained based on the widely used least square (LS) channel estimation, with
\begin{equation}
\widehat{H}_{\rm ba}^{[k]}[m] =\frac{Y^{[k]}[m]}{X^{[k]}[m]}=H_{\rm ba}^{[k]}[m]+\frac{Z[m]}{X^{[k]}[m]}.
\end{equation}
The magnitude of the CSI estimation $\widehat{\bm H}_{\rm ba}^{[k]}$ can be defined as $|\widehat{\bm H}_{\rm ba}^{[k]}|=[|\widehat{H}_{\rm ba}^{[k]}[0]|, |\widehat{H}_{\rm ba}^{[k]}[1]|, \dots, |\widehat{H}_{\rm ba}^{[k]}[M-1]|]$, where $|\widehat{H}_{\rm ba}^{[k]}[m]|$ is the magnitude of the estimated channel coefficient on the $m$-th subcarrier.

\textbf{Threat Model:}
As indicated in~\cite{hoang2024physical}, passive eavesdroppers are very rare in physical layer authentication. In this paper, we consider an active attacker, Mallory, who is located distance $d_{\rm bm}$ away from Bob and impersonates Bob by transmitting packets to Alice in a burst mode. 
The CIR between Mallory and Alice at time $t$ is given as ${\bm h}_{\rm ma}(t)\triangleq[h_{\rm ma}(0,t),h_{\rm ma}(1,t),\dots,h_{\rm ma}(L-1, t)]$. We assume that ${\bm h}_{\rm ma}(t)$ and ${\bm h}_{\rm ba}(t)$ have the same number of channel taps, which is the worst-case since Mallory is least likely to be authenticated.
Mallory knows the wireless protocol that Alice and Bob are using. It also has access to its configuration such as bandwidth, carrier frequency, etc. Mallory aims to spoof Bob by transmitting signals to Alice. We assume that Mallory only performs burst signal attacks, i.e., Mallory does not transmit consecutive packets to Alice. 

\begin{figure}[!t]
\centerline{\includegraphics[width=3.4in]{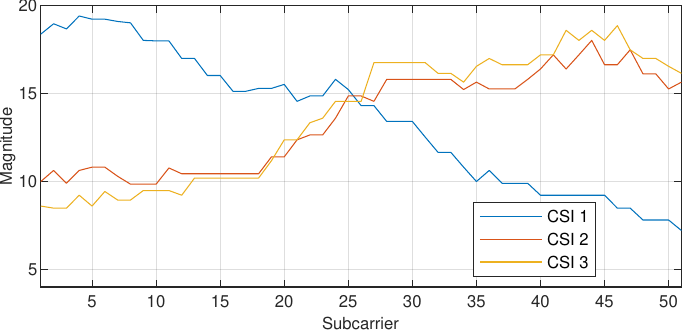}}
\caption{CSI obtained in three different locations using IEEE 802.11n. }
\label{fig:CSIThreeLocations}
\end{figure}

When Alice receives the $(k+1)$-th packet, she will estimate the channel coefficients from its received packet, which can be either from Bob or Mallory. In other words, $\widehat{\bm H}^{[k+1]}$ can be $\widehat{\bm H}_{\rm ba}^{[k+1]}$ or $\widehat{\bm H}_{\rm ma}^{[k+1]}$.
If the packet is from Bob (i.e., $\widehat{\bm H}^{[k+1]}=\widehat{\bm H}_{\rm ba}^{[k+1]}$), the similarity between $\widehat{\bm H}^{[k+1]}$ and $\widehat{\bm H}_{\rm ba}^{[k]}$ will be high, because Bob will not move too far away for practical transmission intervals and typical terminal speeds.
Fig.~\ref{fig:CSIThreeLocations} exemplifies three CSI measurements collected by IEEE 802.11n. The collection sites of CSI $2$ and CSI $3$ are spaced $0.25$ cm and the two CSI estimations are highly similar. 
In contrast, if the packet is transmitted by the attacker Mallory (i.e., $\widehat{\bm H}^{[k+1]}=\widehat{\bm H}_{\rm ma}^{[k+1]}$), $\widehat{\bm H}^{[k+1]}$ will be dissimilar from $\widehat{\bm H}_{\rm ba}^{[k]}$, as exemplified in Fig.~\ref{fig:CSIThreeLocations}
where the distance between the collection site of CSI~1 and the collection sites of CSI $2$ is about $0.5$ m. 
Therefore, by comparing the similarity of CSI measurements between the adjacent two packets, i.e., $\widehat{\bm H}_{\rm ba}^{[k]}$ and $\widehat{\bm H}^{[k+1]}$, Alice will be able to authenticate the identity of the transmitter.

\subsection{Problem Statement}
Correlation is a straightforward approach to quantify the CSI similarity. The fluctuation of CSI is related to the wireless environment and the mobile speed of the device, which means that the time-variant CSI is temporally and spatially correlated. Indeed, the work in~\cite{liu2018authenticating} designs a correlation-based method for device authentication in mobile scenarios, which is used as the benchmark algorithm in this paper.

The Pearson correlation coefficient of the two CSI measurements, $\widehat{\bm H}_{\rm ba}^{[k]}$ and $\widehat{\bm H}^{[k+1]}$, can be defined as
\begin{equation}
\label{eq:pearsoncorrelation}
r({\bm X_1}, {\bm X_2})=\frac{({\bm X_1}-\bar{\bm X_1})^{T}({\bm X_2}-\bar{\bm X_2})}{\| {\bm X_1}-\bar{\bm X_1}\| \|{\bm X_2}-\bar{\bm X_2}\|},
\end{equation}
where ${\bm X_1} = |\widehat{\bm H}_{\rm ba}^{[k]}|$, ${\bm X_2} = |\widehat{\bm H}^{[k+1]}|$, and $\bar{{\bm X}} = 1/M\cdot ({\bm 1}_{1\times M} \cdot {\bm X}) \cdot {\bm 1}_{M\times 1}$.
% \begin{equation}
% r(\widehat{\bm H}_{\rm ba}^{[k]}, \widehat{\bm H}^{[k+1]})=\frac{(\widehat{\bm H}_{\rm ba}^{[k]}-\bar{\bm H}_{\rm ba}^{[k]})^{T}(\widehat{\bm H}^{[k+1]}-\bar{\bm H}^{[k+1]})}{\| \widehat{\bm H}_{\rm ba}^{[k]}-\bar{\bm H}_{\rm ba}^{[k]}\| \|\widehat{\bm H}^{[k+1]}-\bar{\bm H}^{[k+1]}\|},
% \end{equation}
% where $\bar{\bm H}_{\rm ba}^{[k]}=1/M\cdot ({\bm 1}_{1\times M} \cdot \widehat{\bm H}_{\rm ba}^{[k]}) \cdot {\bm 1}_{M\times 1}$ and $\bar{\bm H}^{[k+1]}=1/M\cdot ({\bm 1}_{1\times M} \cdot \widehat{\bm H}^{[k+1]}) \cdot {\bm 1}_{M\times 1}$.
After obtaining the correlation between adjacent CSI measurements, a detection mechanism is given as 
\begin{equation}
\label{eq:pearsonthreshold}
D=\left\{\begin{array}{rl}
    1, & \text{when } r \leq \epsilon_{\rm c}; \\
	0, & \text{when } r > \epsilon_{\rm c},
	\end{array} \right.
\end{equation}
where $D=1$ denotes that there is a rogue device, $D=0$ denotes that there is no rogue devices and a threshold $\epsilon_{\rm c}$ is obtained empirically through experiments.

However, correlation-based authentication relies directly on CSI measurements. When the SNR is low, the quality of CSI measurements degrades, making it less robust to noise. Besides, when Mallory is close to Bob, due to the spatial correlation, ${\bm h}_{\rm ma}(t_{k+1})$ and ${\bm h}_{\rm ba}(t_{k})$ will be highly similar.
The relationship between the distance and the detection capability is not studied yet but is very important.

\section{Proposed Method}
\label{sec:proposed}
\begin{figure}[!t]
\centerline{\includegraphics[width=3.4in]{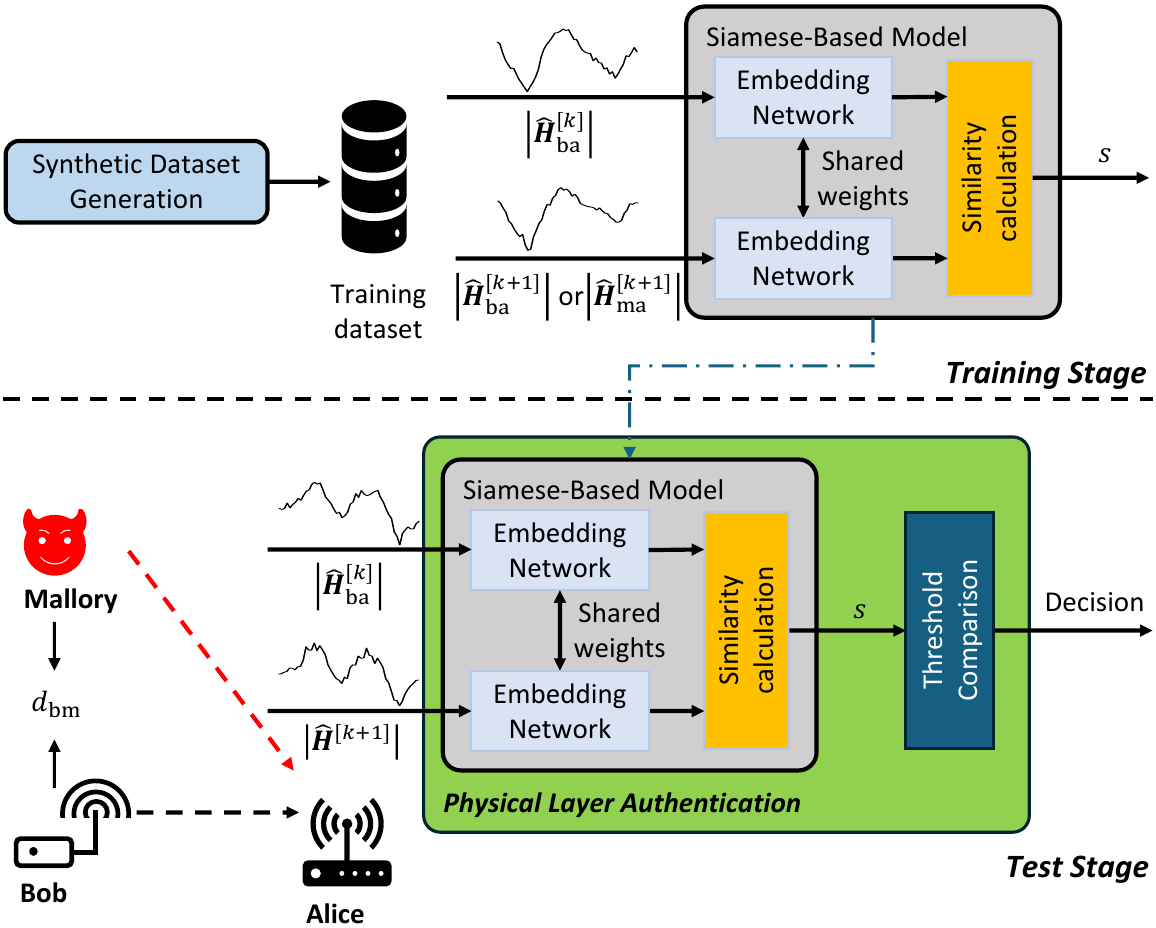}}
\caption{System diagram of the proposed authentication method. }
\label{fig:AuthenticationSystem}
\end{figure}

We propose a deep learning-based method to learn the similarity between a pair of CSI estimations. By incorporating a wide variety of channel models and SNRs into the training dataset, the prior information about the channel environment can be learned, which helps to improve the robustness of the deep learning model.
Therefore, we design a deep learning Siamese network-based approach, enhanced by a synthetic training dataset, as shown in Fig.~\ref{fig:AuthenticationSystem}. The approach consists of the training and test stages.

\subsection{Training Stage} 
For deep learning-based methods, a training dataset $\mathcal{D}_{\text{train}}=\{({\bm U}_{i},V_{i})\}_{i=1}^{|\mathcal{D}|}$ is essential, where $|\mathcal{D}|$ is the cardinality of set $\mathcal{D}_{\rm train}$ and 
\begin{equation}\label{eq:syntheticdataset}
{\bm U}_{i}\triangleq\left\{\begin{array}{rl}
    (\widehat{\bm H}_{\rm ba}^{[k]}, \widehat{\bm H}_{\rm ba}^{[k+1]}), & \text{when } V_{i}=0; \\
	(\widehat{\bm H}_{\rm ba}^{[k]}, \widehat{\bm H}_{\rm ma}^{[k+1]}), & \text{when } V_{i}=1.
	\end{array} \right.
\end{equation}
Assuming the $k$-th packet is transmitted by Bob, we set $V_{i}=0$ when the $(k+1)$-th received packet is from Bob, treating the CSI measurements are highly correlated and $V_{i}=1$ when the $(k+1)$-th received packet is from Mallory, treating them uncorrelated. 
There are currently two ways to obtain a training dataset: experimental collection of CSI measurements or artificial synthesis of CSI estimations. However, collecting a comprehensive dataset from experiments is usually time-consuming and labor-intensive. Therefore, this paper proposes to generate abundant and accurate CSI estimations by simulation as a synthetic training dataset. The details of the synthetic data generation will be introduced in Section~\ref{sec:synthetic}.

Furthermore, the similarity between CSI measurement pairs is the key to device authentication, and the Siamese network is well-suited for learning the similarity between two inputs. Therefore, a Siamese network is exploited. It uses two identical CNN-based embedding networks to work on two different input vectors and computes comparable output values. The goal is to make the Siamese-based model learn a similarity function that measures how similar the two input vectors are and returns a similarity value. In this work, a low score is returned when the input vectors are similar and a high score is returned when the input vectors are different. A contrastive loss function is then calculated.
The details of the Siamese-based model will be introduced in Section~\ref{sec:siamese}.

\subsection{Test Stage} 
In the test stage, the trained model will be used to detect rogue users. When Alice gets two CSI estimations $\widehat{\bm H}_{\rm ba}^{[k]}$ and $\widehat{\bm H}^{[k+1]}$, she will pass the magnitude of them to the trained Siamese network, which will return a score $s$. The score $s$ is compared with a threshold to make the decision:
\begin{equation}
D=\left\{\begin{array}{rl}
        1, & \text{when } s > \epsilon_{\rm s}; \\
	0, & \text{when } s \leq \epsilon_{\rm s},
	\end{array} \right.
\end{equation}
where $D=1$ denotes that there is a rogue device, and $D=0$ denotes that there is no rogue device. The threshold $\epsilon_{\rm s}$ is obtained empirically through experiments.

In the simulation and experimental evaluation, which will be elaborated in Section~\ref{sec:simulation} and Section~\ref{sec:experiment}, respectively, the true positive rate (TPR), false positive rate (FPR), receiver operating characteristics (ROC) and area under the curve (AUC) are used as the performance metrics. Specifically, 
% \begin{equation}
% \text{TPR}=\frac{\text{TP}}{\text{TP+FN}},
% \end{equation}
% where TP is the number of correctly predicted positive samples and FN is the number of actual positive sample incorrectly predicted as negative. And
% \begin{equation}
% \text{FPR}=\frac{\text{FP}}{\text{FP+TN}},
% \end{equation}
% where FP is the number of actual negative samples incorrectly predicted as positive and TN is the number of correctly predicted negative samples.
TPR is defined as the proportion of correctly classified positive samples to the total positive samples, and FPR is defined as the proportion of misclassified negative samples to the total negative samples. 
Based on TPR and FPR, the ROC curve and the AUC can be obtained for performance evaluation. The ROC curve plots TPR versus FPR and is a common performance metric for classification problems under various threshold settings. AUC represents the area under the ROC curve, which tells how well the model is able to discriminate between categories. The higher the AUC, the better the performance for authentication.

\section{Synthetic Dataset Generation}
\label{sec:synthetic}
In this section, the IEEE 802.11n operating in 2.4 GHz carrier frequency and the Rayleigh fading channel are considered. 
We first describe the discrete channel model of WLAN TGn channel, then introduce the synthetic channel generation process, and elaborate on the generation of the synthetic training dataset.

\subsection{WLAN TGn Channel Model}
A set of WLAN channel models is proposed for different environments, as shown in Table~\ref{tab:WLANchannelmodels}. The power delay profile (PDP) of each model is defined based on the cluster modeling approach, where multiple clusters are assigned to models and each cluster is outlined by exponential decay. The specific PDP value of each model can be found in~\cite{Erceg2004IEEEPW}.
\begin{table}[!t]
    \centering
    \caption{WLAN TGn Channel Models~\cite{Erceg2004IEEEPW}.}
    \begin{tabular}{cccc}
         \hline
         Model & RMS Delay (ns)& No. Clusters&Mapped Environment\\
         \hline
         Model B&15&2& Residential apartment\\
         %\hline
         Model C&30&2& Small office\\
         %\hline
         Model D&50&3& Typical office\\
         %\hline
         Model E&100&4& Large office\\
         %\hline
         Model F&150&6& Large space\\
         \hline
    \end{tabular}
    \label{tab:WLANchannelmodels}
\end{table}

In indoor wireless systems, the Doppler spectrum is defined as the Bell shape spectrum, given in the linear scale as~\cite{Erceg2004IEEEPW}
\begin{equation}
S(f)=\frac{\sqrt{A}}{\pi f_{\rm d}} \cdot \frac{1}{1+A(\frac{f}{f_{\rm d}})^2},
\end{equation}
where $A$ is a constant 9 and the Doppler spread $f_{\rm d}$ is
\begin{equation}\label{eq:dopplerspread}
f_{\rm d}=\frac{v_0}{\lambda},
\end{equation}
where $v_0$ is the terminal moving speed and $\lambda$ is the wavelength. Therefore, the autocorrelation function can be calculated as
\begin{equation}
R(\Delta t)=e^{-\frac{2 \pi f_{\rm d}}{\sqrt{A}}\Delta t},
\end{equation}
where $\Delta t$ represents the time interval.

\subsection{Synthetic Channel Generation}
Assuming that $h_{\rm ba}[l] \sim \mathcal{C N}(0, \sigma_{\rm ba}^2(l)), l=0, 1, \cdots, L-1$, and the channel samples of all taps follow the same Doppler spectrum, the channel $h_{\rm ba}^{[k+1]}[l]$ can be written as~\cite{zhang2017on}
\begin{equation}\label{eq:channelcorrelation1}
    h_{\rm ba}^{[k+1]}[l]=R(\Delta t_{k}) h_{\rm ba}^{[k]}[l]+\sqrt{1-R^2(\Delta t_{k})} \omega_{1}[l],
\end{equation}
where $\Delta t_{k}=t_{k+1}-t_{k}$ denotes the transmission interval, $R(\Delta t_{k})$ denotes the autocorrelation coefficient of channel $h_{\rm ba}[l]$ and the random component $\omega_{1}[l] \sim \mathcal{C N}(0, \sigma_{\rm ba}^2(l))$. 
The corresponding CFR can be expressed as 
\begin{align}
& H_{\rm ba}^{[k+1]}[m]=\sum_{l=0}^{L-1} h_{\rm ba}^{[k+1]}[l] e^{-j 2 \pi m l/M} \nonumber\\
& =R(\Delta t_{k}) H_{\rm ba}^{[k]}[m]+\sqrt{1-R^2(\Delta t_{k})} \sum_{l=0}^{L-1} \omega_{1}[l] e^{-j 2 \pi m l/M}\nonumber\\
& =R(\Delta t_{k}) H_{\rm ba}^{[k]}[m]+\sqrt{1-R^2(\Delta t_{k})} \Omega_{1}[m].
\end{align}

Given the distance between Bob and Mallory $d_{\rm bm}$, it can be derived from \eqref{eq:dopplerspread} that the equivalent time interval between Bob and Mallory is
\begin{equation}
    \Delta t_{\rm bm}=\frac{d_{\rm bm}}{v_0}=\frac{d_{\rm bm}}{f_{\rm d}\lambda}.
\end{equation}
The channel $h_{\rm ma}^{[k+1]}[l]$ can be written as~\cite{michalopoulos2012amplify}
\begin{equation}\label{eq:channelcorrelation2}
    h_{\rm ma}^{[k+1]}[l]=\frac{1}{\sqrt{\Theta}}(\rho(d_{\rm bm}) h_{\rm ba}^{[k+1]}[l]+\sqrt{1-\rho^2(d_{\rm bm})} \omega_{2}[l]),
\end{equation}
where $\rho(d_{\rm bm})=R(\Delta t_{\rm bm})$ denotes the correlation coefficient between $h_{\rm ba}^{[k+1]}[l]$ and $h_{\rm ma}^{[k+1]}[l]$ spaced by distance $d_{\rm bm}$ and the random component $\omega_{2}[l] \sim \mathcal{C N}(0, \sigma_{\rm ba}^2(l))$ and $\Theta=\sigma_{\rm ba}^2(l) / \sigma_{\rm ma}^2(l)$, where we assume that $\Theta$ is a constant for different channel tap.
The corresponding CFR can be expressed as 
\begin{align}
& H_{\rm ma}^{[k+1]}[m]=\sum_{l=0}^{L-1} h_{\rm ma}^{[k+1]}[l] e^{-j 2 \pi m l/M} \nonumber\\
& =\frac{\rho(d_{\rm bm})}{\sqrt{\Theta}} H_{\rm ba}^{[k+1]}[m]+\frac{\sqrt{1-\rho^2(d_{\rm bm})}}{\sqrt{\Theta}} \sum_{l=0}^{L-1} \omega_{2}[l] e^{-j 2 \pi m l/M}\nonumber\\
& =\frac{\rho(d_{\rm bm})}{\sqrt{\Theta}} H_{\rm ba}^{[k+1]}[m]+\frac{\sqrt{1-\rho^2(d_{\rm bm})}}{\sqrt{\Theta}} \Omega_{2}[m].
\end{align}

\subsection{Synthetic Training Dataset Generation Using MATLAB}\label{sec:synthetic_training}
The synthetic dataset $\mathcal{D}_{\rm train}^{\rm S}=\{({\bm U}_{i}^{\rm S},V_{i}^{\rm S})\}_{i=1}^{|\mathcal{D}_{\rm S}|}$ contains synthetic CSI pairs, where $|\mathcal{D}_{\rm S}|$ is the cardinality of set $\mathcal{D}_{\rm train}^{\rm S}$.
Specifically, the channel ${\bm h}_{\rm ba}(t_{k})$ is generated based on IEEE 802.11 TGn channel models by using MATLAB WLAN toolbox\footnote{https://www.mathworks.com/help/wlan/ref/wlantgnchannel-system-object.html}. The channel ${\bm h}_{\rm ba}(t_{k+1})$ and ${\bm h}_{\rm ma}(t_{k+1})$ are generated based on \eqref{eq:channelcorrelation1} and \eqref{eq:channelcorrelation2}, respectively. A WLAN Non-HT format waveform is generated and is filtered by channel ${\bm h}_{\rm ba}(t_{k})$, ${\bm h}_{\rm ba}(t_{k+1})$ and ${\bm h}_{\rm ma}(t_{k+1})$, respectively. Then AWGN is added. At the receiver, the LTF is extracted and CSI estimations, i.e., $\widehat{\bm H}_{\rm ba}^{[k]}$, $\widehat{\bm H}_{\rm ba}^{[k+1]}$, and $\widehat{\bm H}_{\rm ma}^{[k+1]}$, are obtained. 
We can then construct the synthetic dataset based on~\eqref{eq:syntheticdataset}.

\begin{figure}[!t]
    \centerline{\includegraphics[width=3.4in]{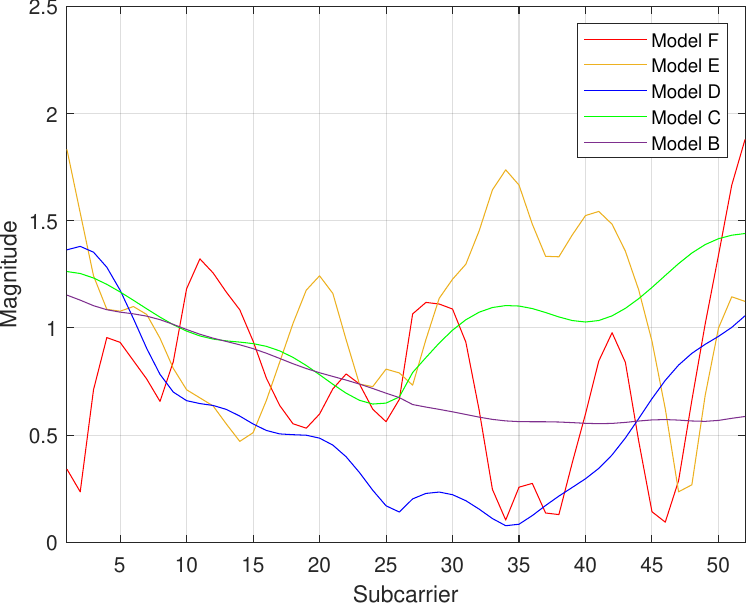}}
    \caption{CSI of models B-F without noise.}
    \label{fig:CSIofModelBCDEF}
\end{figure}

In order to enable the synthetic training dataset to align with the practical scenarios and be generalizable, we carefully tuned the following parameters. 
\begin{itemize}
    \item Different multipath environments: WLAN TGn channel models B-F. Fig.~\ref{fig:CSIofModelBCDEF} exemplifies the CSI of WLAN TGn channel models B-F without noise using MATLAB simulation. It can be observed that a large RMS delay causes severe fluctuations in CSI magnitudes, which makes the channel have strong frequency selectivity. 
    \item SNR: 5, 8, 10, 12, 13, 14, 15, 16, 17, 18, 19, 20 and 50~dB. 
    \item Transmission interval $\Delta t_{k}$: 3~ms.
    \item The moving speed $v_0$: 1~m/s. 
    \item Distance between Bob and Mallory $d_{\rm bm}$: 0.25, 0.5, 0.75, 1, 1.5, 2 and 3 wavelengths.
\end{itemize}
The training synthetic dataset will include all the configurations iterating different combinations of the above parameters.

\section{Siamese-based CSI Authentication}
\label{sec:siamese}
\begin{figure}[!t]
\centerline{\includegraphics[width=3.4in]{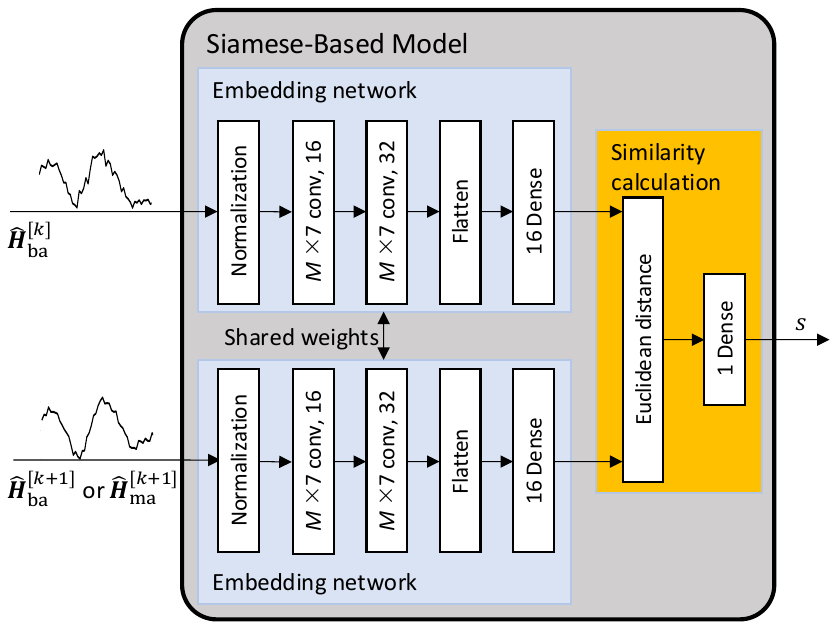}}
\caption{The Siamese-based model.}
\label{fig:SiameseModel}
\end{figure}

The CNN-based Siamese network is employed to learn the similarity between channel estimations.% in synthetic training dataset $\mathcal{D}_{\rm train}^{\rm S}$. 
The architecture of the proposed Siamese model is illustrated in Fig.~\ref{fig:SiameseModel}. The inputs are the magnitudes of two channel estimations, i.e., $|\widehat{\bm H}_{\rm ba}^{[k]}|$ and $|\widehat{\bm H}_{\rm ba}^{[k+1]}|$, or $|\widehat{\bm H}_{\rm ba}^{[k]}|$ and $|\widehat{\bm H}_{\rm ma}^{[k+s1]}|$.
The Siamese-based model consists of two twin embedding networks and a similarity calculation module. 

The CNN-based embedding network can be regarded as a feature extractor, which is composed of a normalization layer, two convolution layers, a flatten layer, and a dense layer. 
The input CSI estimations are normalized by min-max normalization. The convolution layers use 16 $M\times 7$ and 32 $M\times 7$ filters, respectively. Both of the convolution layers are activated by the ReLU function and padding is used. Two identical embedding networks act on two input CSI estimations to extract features, respectively. 
The initial parameters of the twin embedding networks are identical, so that features of the two input CSI estimations are extracted equivalently. 

Each embedding network returns a $16\times 1$ feature vector, and the Euclidean distance between them is calculated. A score $s$ that measures the similarity between two CSI estimations is generated in the last dense layer with the sigmoid activation function, which produces a score $s$ between 0 and 1. 

The contrastive loss function is adopted to train the Siamese model, defined as
\begin{equation}
\mathcal{L} =V_{i}\{\max (0, \eta-s)\}^2+(1-V_{i}) s^2,
\end{equation}
where $V_{i}$ denotes the label of the input CSI pair and $\eta=1$ is a margin parameter. It indicates that a pair of similar CSI estimations will make $s$ approach 0, while a pair of dissimilar CSI estimations will make $s$ approach 1.

The Siamese-based model architecture is built with Python 3.8 in Tensorflow, and the network is trained on 2 NVIDIA Tesla V100 GPUs using the RMSprop optimizer with a learning rate of 0.001 and a batch size of 32. 
% The synthetic training dataset is randomly partitioned into $90\%$ training data and $10\%$ validation data.

\section{Simulation Evaluation}
\label{sec:simulation}
In this section, the simulation test dataset generation is first described, and then the benchmark algorithms are described, and eventually the performance of the proposed CNN-based Siamese model, FCN-bsed Siamese model and correlation-based method in simulation is evaluated and compared.

\subsection{Simulation Test Dataset Generation}
The simulation test dataset is generated in the same way as the synthetic training dataset introduced in Section \ref{sec:synthetic_training} with parameter configurations as follows.
\begin{itemize}
    \item Different multipath environments: WLAN TGn channel models B-F.
    \item SNR: 0, 2, 4, 6, 8, 10, 12, 14, 16, 18, 20 and 22~dB. 
    \item Transmission interval $\Delta t_{k}$: 3, 6, 9, 12, 15, 18, 21, 24 and 27~ms.
    \item The moving speed $v_0$: 1~m/s. 
    \item Distance between Bob and Mallory $d_{\rm bm}$: 0.25, 0.5, 0.75, 1, 1.5, 2 and 3 wavelengths.
\end{itemize}
It is worth noting that unlike the training dataset, which contains all the different combinations of the parameter settings, the test datasets are generated based on each parameter setting independently, to represent a particular test scenario. 
% For example, we generated a test scenario with WLAN TGn channel model B, $\text{SNR} = 0$~dB, $\Delta t_k = 3$~ms, $v_0 = 1$~m/s, and $d_{\rm bm} = 0.25$~wavelength.

\subsection{Benchmark Algorithms}
Pearson correlation-based method in~\cite{liu2018authenticating} and FCN-based Siamese method in~\cite{zhang2025enhancing} are used as benchmark algorithms. The Pearson correlation is calculated as~\eqref{eq:pearsoncorrelation} and the authentication is performed based on an empirically determined threshold, as shown in~\eqref{eq:pearsonthreshold}. The FCN-based Siamese method employs an embedding network with four fully connected layers to extract features from the two input CSI estimations. These layers have 256, 512, 256, and 16 neurons, respectively, with ReLU as the activation function for each.

\subsection{Simulation Results}
% Fig.~\ref{fig:ROC_syn} depicts the ROC curves of the Siamese-based and correlation-based methods for the simulation test dataset. It can be seen that for both methods, the stronger channel frequency selectivity can produce higher TPR and lower FPR. Thanks to the prior knowledge of the channel learned by the Siamese model, the Siamese-based method outperforms the correlation-based method in all WLAN TGn channel models.
% \begin{figure}[!t]
% \centerline{\includegraphics[width=3.4in]{figures/ROC_syn.pdf}}
% \caption{ROC curves of the Siamese-based and correlation-based methods on the simulation test dataset with $\text{SNR}=10$~dB, $\Delta t_k=3$~ms, $d_{\rm bm}/\lambda=0.25$, and $v_0=1$~m/s.}
% \label{fig:ROC_syn}
% \end{figure}

\subsubsection{Comparison with Benchmark}
\begin{figure}[!t]
\centerline{\includegraphics[width=3.4in]{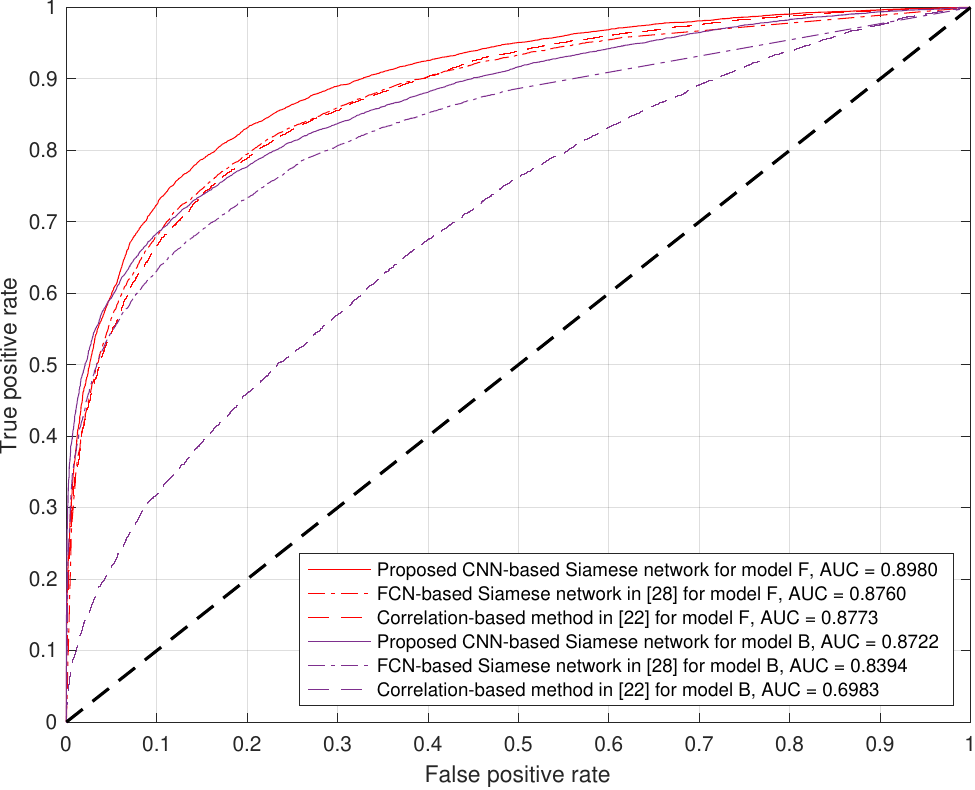}}
\caption{ROC curves of the proposed CNN-based Siamese, FCN-based Siamese and correlation-based method on the simulation test dataset with WLAN TGn channel model B and model F, $\text{SNR}=0$~dB, $\Delta t_k=3$~ms, $d_{\rm bm}/\lambda=0.25$, and $v_0=1$~m/s.}
\label{fig:ROC_syn}
\end{figure}

Fig.~\ref{fig:ROC_syn} depicts the ROC curves of the CNN-based Siamese, FCN-based Siamese and correlation-based method for the simulation test dataset with WLAN TGn channel model B and model F. 

Thanks to the prior knowledge of the channel learned in the training stage, the CNN-based Siamese and FCN-based Siamese models outperform the correlation-based method. Additionally, due to the stronger feature extraction performance of CNN compared to FCN, CNN-based Siamese network has an advantage over FCN-based Siamese network. 

Table~\ref{tab:comparisonSiamese} reflects the computational overhead between CNN-based Siamese and FCN-bsed Siamese method in terms of parameter count and floating point operations (FLOPs). The parameter count refers to the total number of trainable parameters, including weights and biases, across all layers of the model. These parameters are updated during training to minimize the loss function. It can be obtained using the built-in model.summary() tool in TensorFlow.
FLOPs represents the total number of floating point operations required for a single forward pass through the network. This can be calculated using the built-in tf.compat.v1.profiler() tool in TensorFlow. It can be observed that the CNN-based Siamese network has a lower computational overhead compared to the FCN-based Siamese network introduced in~\cite{zhang2025enhancing}.

\begin{table}[!t]
    \centering
    \caption{Comparison of Computational Requirements.}
    \begin{tabular}{ccc}
         \hline
         & Parameter Count& FLOPs\\
         \hline
         FCN-based Siamese network in~\cite{zhang2025enhancing}&282,641&1,435,645\\
         CNN-based Siamese network&29,972&1,175,697\\
         \hline
    \end{tabular}
    \label{tab:comparisonSiamese}
\end{table}

\subsubsection{Evaluation Under Different Settings}
\begin{figure}[!t]
\centering
\subfloat[Siamese-based method.]{\includegraphics[width=3.4in]{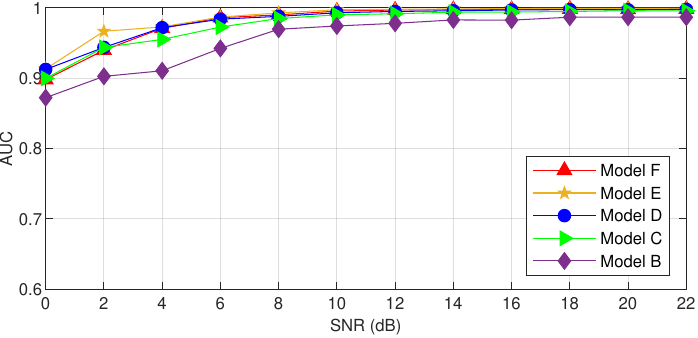}}

\subfloat[Correlation-based method.]{\includegraphics[width=3.4in]{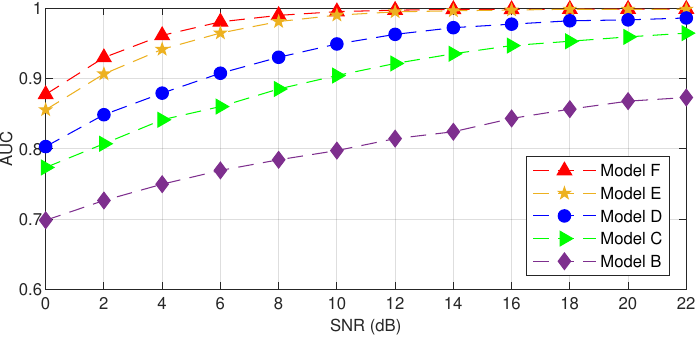}}
\caption{The AUC versus SNR on the simulation test dataset with $d_{\rm bm}/\lambda=0.25$, $\Delta t_{k}=3$ ms and $v_0=1$ m/s.}
\label{fig:AUCvsSNR_simulation}
\end{figure}

Fig.~\ref{fig:AUCvsSNR_simulation} shows the AUC of the CNN-based Siamese network and correlation-based method versus SNR. It can be seen that the authentication performance of both methods gets better with the increasing SNR since the LS channel estimation will be more accurate with higher SNR. The Siamese-based method performs better than the correlation-based method on all SNRs and it works well in all channel models, which means that the CNN-based Siamese network has strong generalization performance for different channel environments. The reason is that by training the Siamese network, the feature extractor learns how to extract feature information from noisy CSI estimations and gets more robust to noise.

\begin{figure}[!t]
\centering
\subfloat[Siamese-based method.]{\includegraphics[width=3.4in]{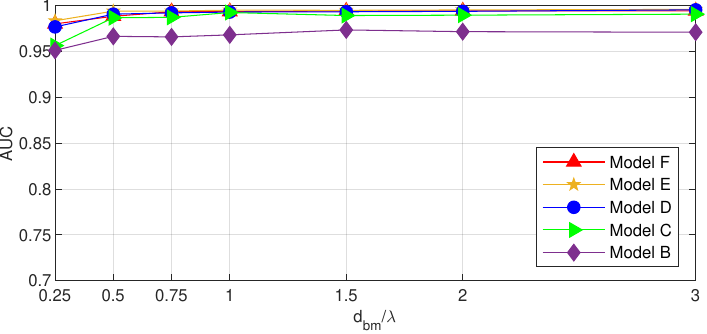}}

\subfloat[Correlation-based method.]{\includegraphics[width=3.4in]{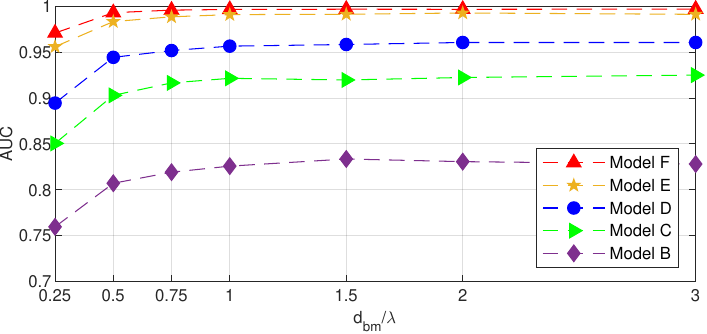}}
\caption{The AUC versus $d_{\rm bm}/\lambda$ on the simulation test dataset with $\text{SNR}=5$~dB, $\Delta t_{k}=3$ ms and $v_0=1$~m/s.}
\label{fig:AUCvsD_simulation}
\end{figure}

Fig.~\ref{fig:AUCvsD_simulation} shows the AUC of the CNN-based Siamese network and the correlation-based method versus the distance between Bob and Mallory normalized by wavelength $d_{\rm bm}/\lambda$. It can be illustrated that with the distance between Bob and Mallory increasing, Alice has a better performance, i.e., higher AUC, detecting rogue devices. For the correlation-based method, the weak frequency selectivity makes it difficult for the receiver to detect rogue devices. However, as for the CNN-based Siamese network, it works very well for all channel models. The reason is that by training the Siamese network, the feature extractor learns how to extract feature information from the channel with weak frequency selectivity.

\begin{figure}[!t]
\centering
\subfloat[Siamese-based method.]{\includegraphics[width=3.4in]{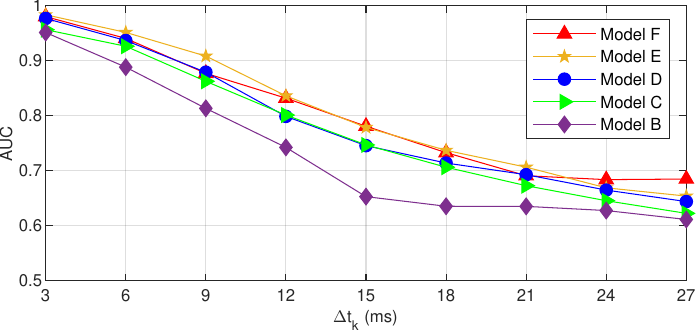}}

\subfloat[Correlation-based method.]{\includegraphics[width=3.4in]{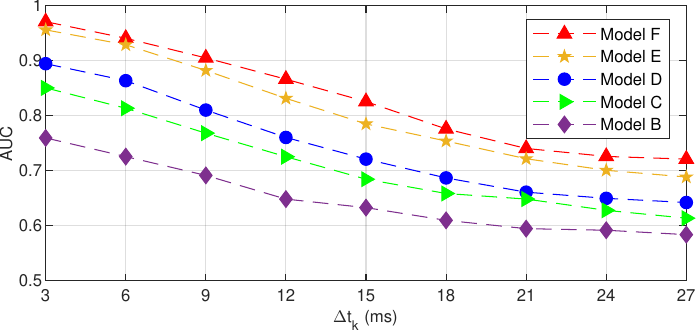}}
\caption{The AUC versus $\Delta t_{k}$ on the simulation test dataset with $d_{\rm bm}/\lambda=0.25$, $\text{SNR}=5$ dB and $v_0=1$ m/s.}
\label{fig:AUCvsT_simulation}
\end{figure}

Fig.~\ref{fig:AUCvsT_simulation} shows the AUC of the CNN-based Siamese network and correlation-based method versus the transmission interval. It can be observed that the AUC decreases as the transmission interval $\Delta t_{k}$ increases because with a certain moving speed, the larger the transmission interval, the lower the correlation between the CSI estimations obtained from two consecutive legitimate packets, and the more difficult it is for Alice to detect the rogue device. Moreover, it also shows that with $\Delta t_{k}$ less than $15$~ms, the CNN-based Siamese network produces higher AUC than the correlation-based method for all channel models, which means that the CNN-based Siamese network has a higher tolerance for different transmission intervals.

\section{Experimental Evaluation}
\label{sec:experiment}
We have verified the effectiveness of the CNN-based Siamese network in the simulation environment in Section~\ref{sec:simulation}. However, the simulation environment may not fully represent the practical environment. In addition, the synthetic training dataset needs to be compared with the experimental training dataset to prove the reliability of using the synthetic dataset. Therefore, experimental evaluation is also conducted. In this section, the experiment setup is first elaborated, and then the performance of the proposed synthetic training dataset is evaluated and compared, and eventually the threshold selection is discussed.

\subsection{Experiment Setup}
\subsubsection{Device Configuration}
\begin{figure}[!t]
\centerline{\includegraphics[width=3in]{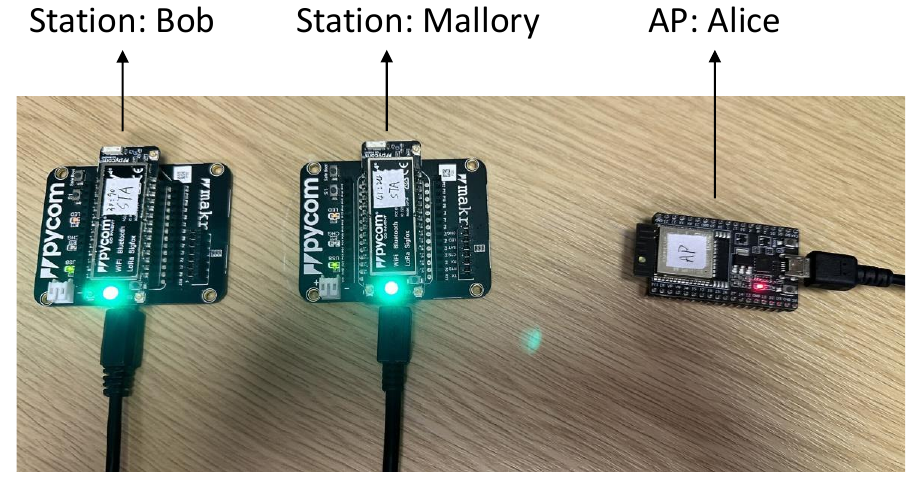}}
\caption{Experiment devices: (i) ESP32 in the WiFi AP mode is used as Alice, and (ii) two LoPy4 boards in the WiFi station mode are used as Bob and Mallory.}
\label{fig:ExperimentDevices}
\end{figure}

Alice is configured as a WiFi access point (AP) while Bob and Mallory are treated as user stations, as shown in Fig.~\ref{fig:ExperimentDevices}. An ESP32 kit with integrated WiFi connectivity is adopted as Alice. The collection of CSI from the ESP32 microcontroller is achieved by ESP32 CSI Toolkit\footnote{https://github.com/StevenMHernandez/ESP32-CSI-Tool}, which can provide information about the working mode, MAC address of the transmitter, RSSI, noise floor, time stamp of the received signal, and CSI. The ESP32 board transfers the collected data to a PC via a USB cable. In addition, two LoPy4 development boards\footnote{https://development.pycom.io/tutorials/networks/wlan/} operating under the WiFi station mode are used as Bob and Mallory. All boards support IEEE 802.11n with a configuration of $20$ MHz bandwidth and $2.4$ GHz carrier frequency. 

\subsubsection{Experiment Scenarios}
\begin{figure}[!t]
\centering
\subfloat[]{\includegraphics[width=3.2in]{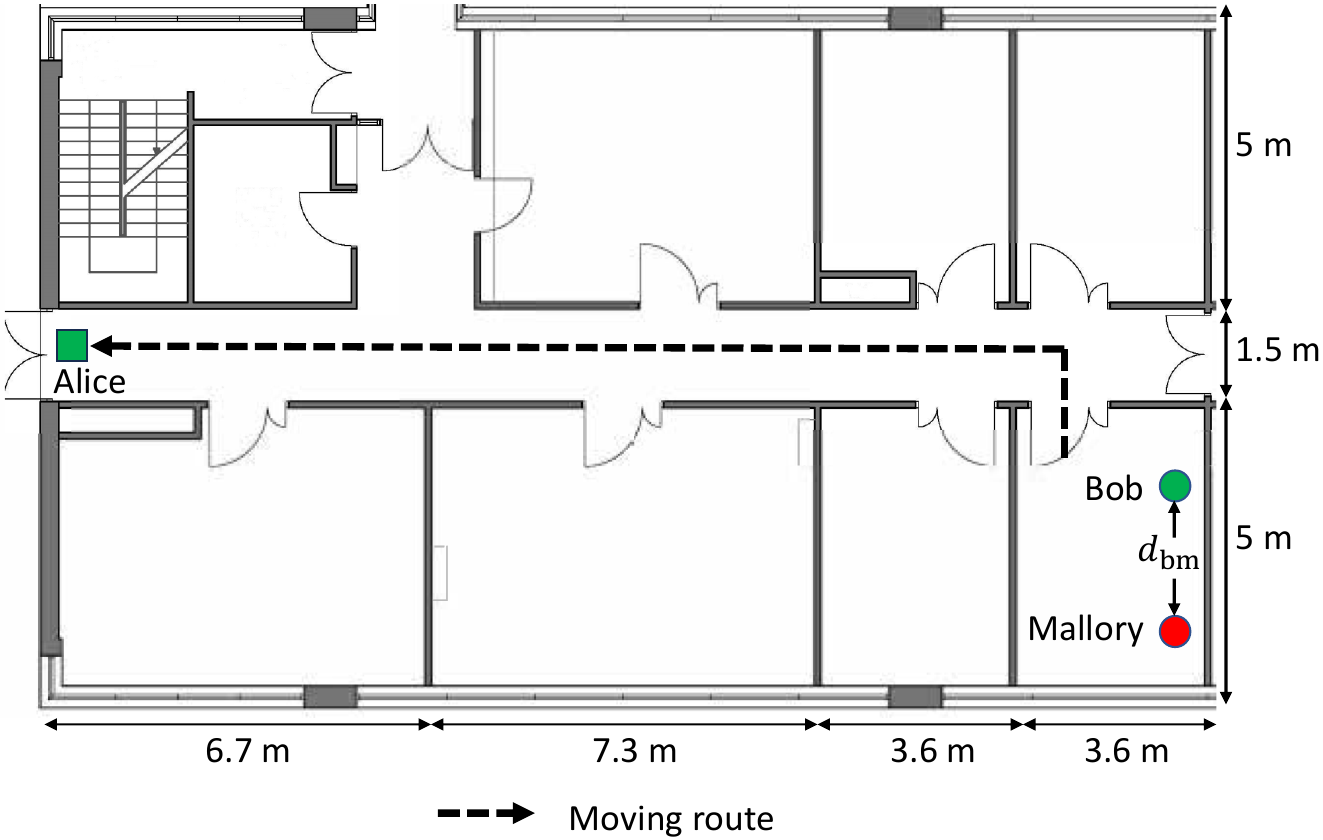}}

\subfloat[]{\includegraphics[width=3.4in]{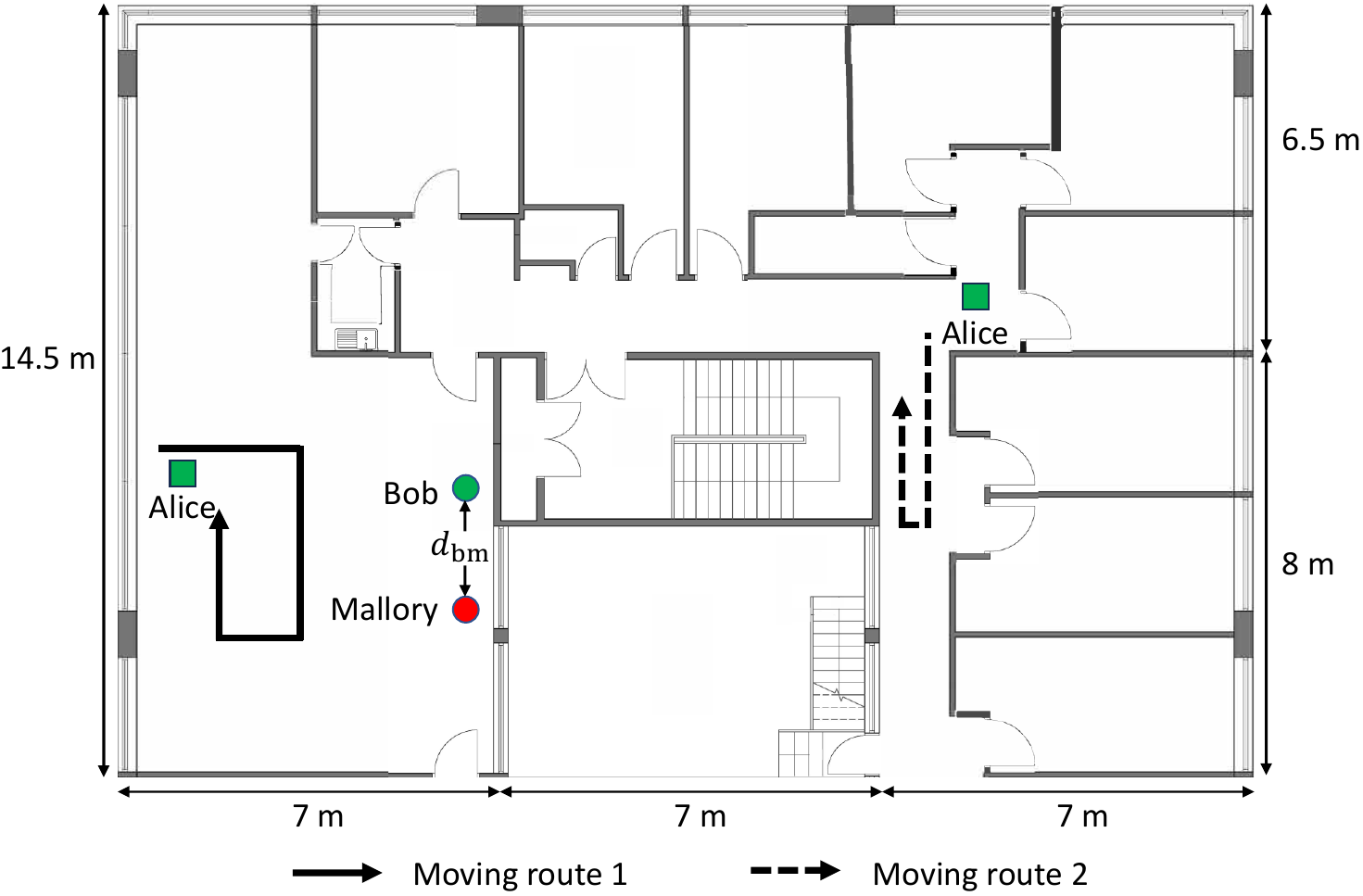}}

\caption{Experiment scenarios. (a) Scenario I: corridor A. (b) Scenario II: office (Moving route 1), and Scenario III: corridor B (Moving route 2). }
\label{fig:TestScenarios}
\end{figure}

To evaluate the performance of the proposed scheme, we consider four different scenarios:
\begin{itemize}
    \item Scenario I: movement in a corridor A, as shown in Fig.~\ref{fig:TestScenarios}(a).
    \item Scenario II: movement in an office, as shown in Fig.~\ref{fig:TestScenarios}(b) (moving route 1).
    \item Scenario III, movement in a corridor B, as shown in Fig.~\ref{fig:TestScenarios}(b) (moving route 2).
    \item Scenario IV: movement in a residential apartment (floor plan not shown).
\end{itemize}
The experiments of scenario I and scenarios II \& III were carried out on the second floor and the sixth floor of the Department of Electrical Engineering and Electronics, the University of Liverpool, UK, respectively.

It is worth noting that the datasets used for training and testing were independently collected in all scenarios.

\subsubsection{Experimental Training Dataset Collection}
Although we have generated the synthetic training dataset, we deliberately collected experimental training datasets for comparison.
Training datasets were collected between Alice and Bob in all scenarios. In the experiments, Bob sent packets continuously from a fixed position at a time interval of $0.01$\,s. Alice moved at a speed of approximately $0.25$\,m/s while collecting measurement data using the ESP32 CSI toolkit.

Due to the movement by Alice, measurements taken at vastly different time instants correspond to those taken at widely separated locations, which corresponds to different devices. Therefore, in our experimental training dataset $\mathcal{D}_{\rm train}^{\rm E}=\{({\bm U}_{i}^{\rm E},V_{i}^{\rm E})\}_{i=1}^{|\mathcal{D}_{\rm E}|}$, where each input ${\bm U}_{i}^{\rm E}\triangleq (\widehat{\bm H}^{[k]},\widehat{\bm H}^{[k+\Delta k]})$ corresponds to a pair of channel measurements separated by $\Delta k$ packets, we set $V_{i}^{\rm E}=0$, when $\Delta k=1$, treating the CSI measurements as those coming from the same device, and $V_{i}^{\rm E}=1$, when $\Delta k = 100$, treating them as those coming from different devices. We collected $5145$, $5037$, $5077$ and $4240$ CSI measurements in scenario I, scenario II, scenario III and scenario IV, respectively.

\subsubsection{Experimental Test Dataset Collection}
In all scenarios, the positions of Bob and Mallory were fixed and separated by distance $d_{\rm bm}$. Bob and Mallory continuously transmitted signals to Alice at time intervals of $0.01$\,s and $0.1$\,s, respectively. The distance between Bob and Mallory $d_{\rm bm}$ can take on the values of $3$, $6$, $9$, $12$, $18$, $24$ and $36$ cm. By operating at 2.4 GHz, where the signal wavelength is about $12$~cm, the distances considered above correspond to $0.25$, $0.5$, $0.75$, $1$, $1.5$, $2$ and $3$ wavelengths. 

Alice moved at a speed of about $0.25$ m/s in all scenarios and received signals from both Bob and Mallory. 
Moreover, since Alice is able to extract the transmitter's MAC address, we can utilize it to generate the ground truth identity of the received packets. It is worthwhile to note that, while the AP is usually fixed in practice, we consider the movement of the AP rather than the movement of user stations in order to control and adjust the distance between Bob and Mallory. Thanks to channel reciprocity, this setup is equivalent to fixed AP and mobile user stations.

\subsection{Experiment Results}
\begin{figure}[!t]
\centering
\subfloat[Training dataset in scenario I.]{\includegraphics[width=1.7in]{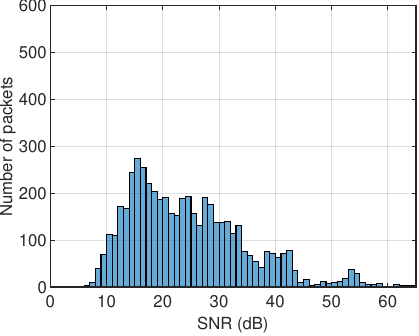}}
\subfloat[Test dataset in scenario I.]{\includegraphics[width=1.7in]{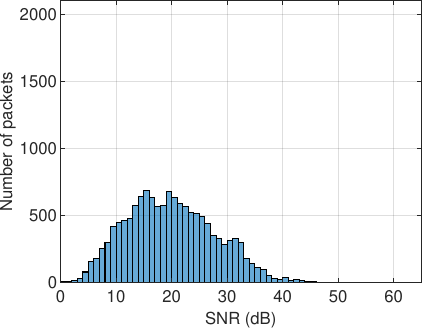}}

\subfloat[Training dataset in scenario II.]{\includegraphics[width=1.7in]{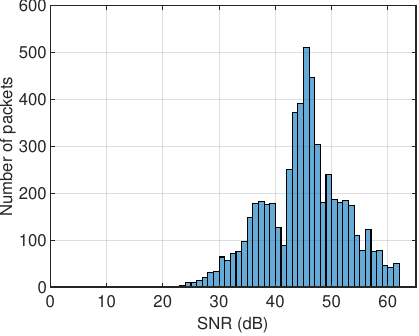}}
\subfloat[Test dataset in scenario II.]{\includegraphics[width=1.7in]{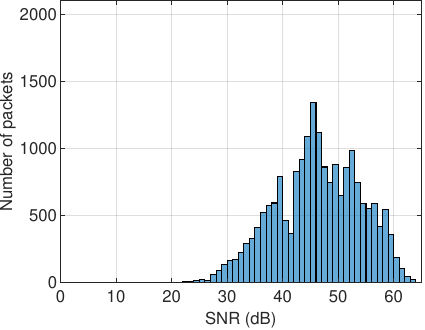}}

\subfloat[Training dataset in scenario III.]{\includegraphics[width=1.7in]{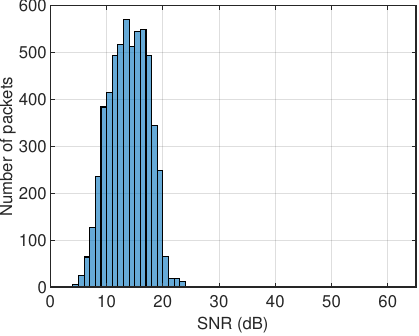}}
\subfloat[Test dataset in scenario III.]{\includegraphics[width=1.7in]{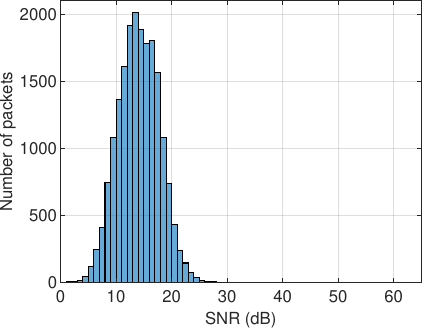}}

\subfloat[Training dataset in scenario IV.]{\includegraphics[width=1.7in]{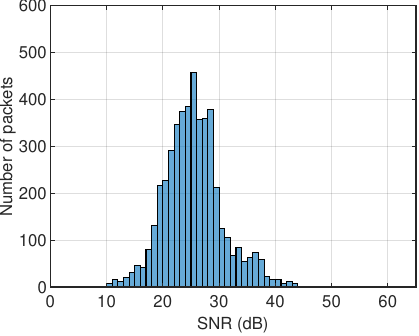}}
\subfloat[Test dataset in scenario IV.]{\includegraphics[width=1.7in]{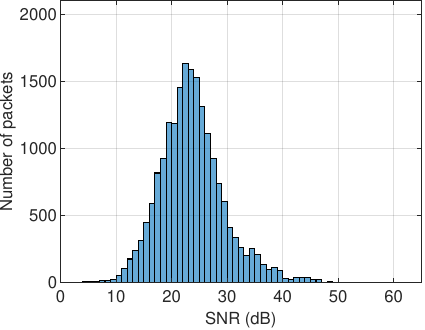}}
\caption{The SNR distribution of the training datasets and test datasets collected in different scenarios.}
\label{fig:SNRdistribution}
\end{figure}

Fig.~\ref{fig:SNRdistribution} shows the SNR distribution of the packets collected in scenarios I-IV for both training and test datasets. It can be seen that scenario II and scenario III are related to relatively high and low SNR environments, respectively. As shown in Fig.~\ref{fig:TestScenarios}, in scenario II, there were always LOS transmissions between Alice and Bob/Mallory, and their distance was short, up to $7$~m.
In contrast, there was only NLOS transmissions available in scenario III and the distance was as large as $15$~m. 
Scenario I and scenario IV represent the medium SNR environment.

\begin{figure}[!t]
\centerline{\includegraphics[width=3.4in]{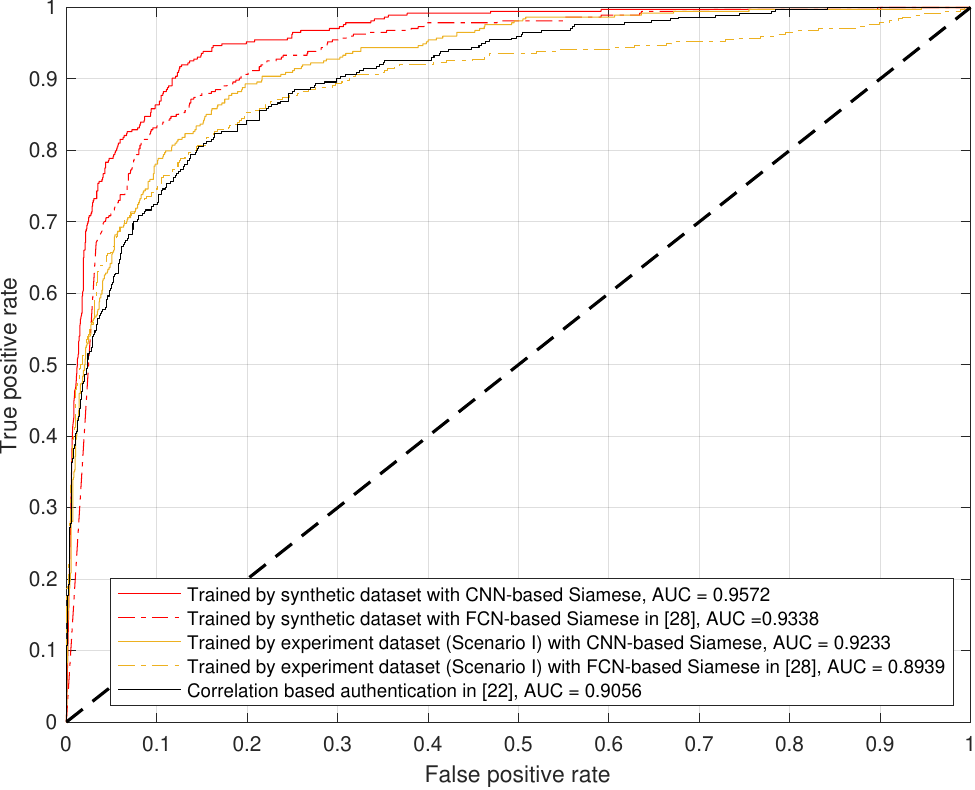}}
\caption{ROC curves of the CNN-based Siamese, FCN-based Siamese and correlation-based method on the test dataset collected in the scenario I (corridor A) with $d_{\rm bm}/\lambda=1$.}
\label{fig:ROC_synVSexp}
\end{figure}

Fig.~\ref{fig:ROC_synVSexp} shows the ROC curves of our proposed CNN-based Siamese, FCN-based Siamese and correlation-based method for the scenario I (corridor A). CNN-based Siamese network and FCN-based Siamese network are trained on the synthetic training dataset and the experimental training dataset collected in scenario I (corridor A), respectively. It can be observed that the CNN-based Siamese network produces 0.03 higher AUC than FCN-based Siamese network and 0.05 higher AUC than the correlation-based method, respectively. That is, for each threshold, the CNN-based Siamese network has a higher TPR and a lower FPR than the FCN-based Siamese network and correlation-based method. More importantly, the Siamese model trained on the synthetic training dataset has the potential to yield higher AUC than the one trained on the experimental training dataset, which demonstrates the advantages of the synthetic dataset.

\begin{figure}[!t]
\centering
\subfloat[Test dataset collected in scenario I.]{\includegraphics[width=3.4in]{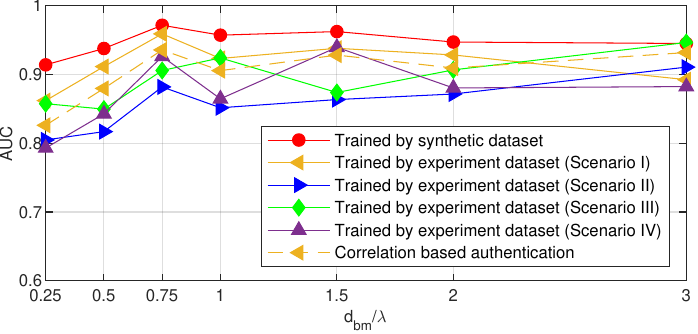}}

\subfloat[Test dataset collected in scenario II.]{\includegraphics[width=3.4in]{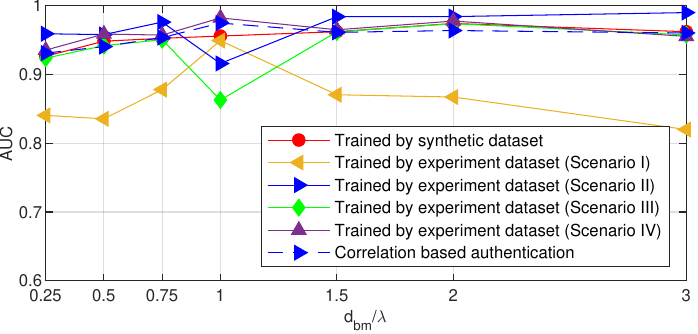}}

\subfloat[Test dataset collected in scenario III.]{\includegraphics[width=3.4in]{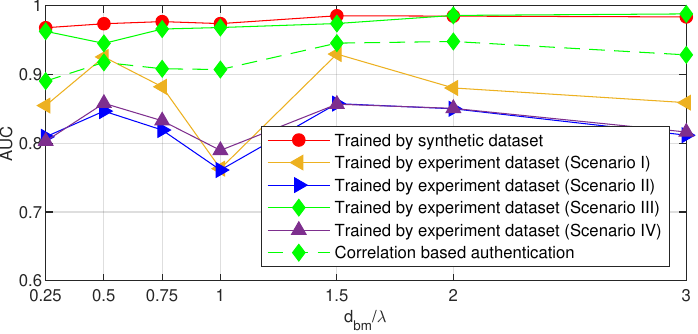}}

\subfloat[Test dataset collected in scenario IV.]{\includegraphics[width=3.4in]{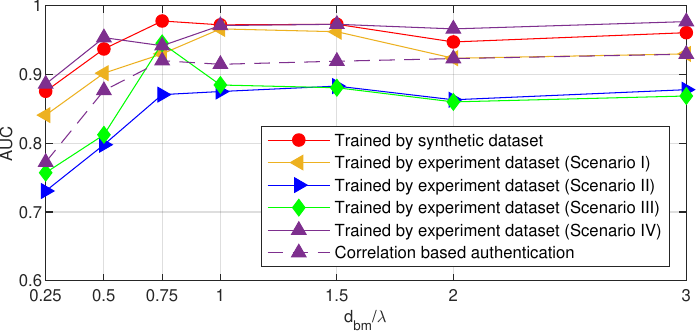}}
\caption{The AUC versus the distance between Bob and Mallory normalized by wavelength $d_{\rm bm}/\lambda$ on various training datasets and test datasets.}
\label{fig:AUC_synVSexp}
\end{figure}

In Fig.~\ref{fig:AUC_synVSexp}, we show the AUC of the CNN-based Siamese network and correlation-based method versus the distance between Bob and Mallory normalized by wavelength $d_{\rm bm}/\lambda$ for all scenarios. The CNN-based Siamese models are trained on the synthetic training dataset and experimental training datasets collected in scenarios I-IV, respectively. We then evaluated these trained models against test datasets collected in different scenarios.
We observe that in all scenarios, our CNN-based Siamese network obtains an average of 0.06 gain of AUC than the correlation-based method. 
Moreover, the synthetic training dataset potentially outperforms the experimental training dataset. 
Taking Fig.~\ref{fig:AUC_synVSexp}(a) as an example. 
When evaluated against the test dataset collected from scenario I, the training data collected in scenario I has better performance than the training datasets collected in other scenarios. As can be observed in Fig.~\ref{fig:SNRdistribution}, the training and test datasets collected in scenario I have similar SNR distributions. Other datasets have deviated SNR distributions, which worsens the generalization. 
For each test scenario, the synthetic training dataset has a comparable performance with the experimental training dataset collected in the matched scenario, which demonstrates good generalization performance across different scenarios.

\begin{figure}[!t]
\centerline{\includegraphics[width=3.4in]{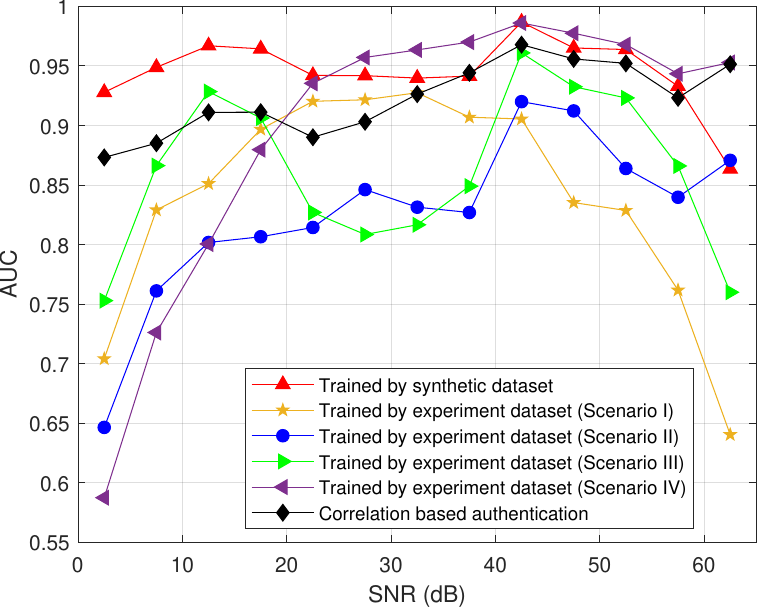}}
\caption{The AUC versus SNR. }
\label{fig:AUCvsSNR_experiment}
\end{figure}

In Fig.~\ref{fig:AUCvsSNR_experiment}, we show the AUC of the CNN-based Siamese network and correlation-based method versus SNR. Specifically, the CSI measurements collected in all scenarios were mixed together, and then divided into 13 groups with a 5 dB interval length based on the SNR of the collected packets (ranging from 0 dB to 65 dB). The SNR of each group of CSI measurements is represented by the median SNR of the interval and each group is tested independently. It can be seen that with SNR lower than $20$ dB, the synthetic dataset always produces an AUC 0.05 higher than the correlation-based method. Moreover, the synthetic dataset performs well in all SNR groups, but the experimental datasets collected in scenarios I-IV perform well only in the corresponding SNR groups. Taking scenario I as an example, since the SNR of the packets collected in scenario I is mainly distributed between $10$ dB and $40$ dB, the experimental training dataset does not perform well with SNR lower than $10$ dB or higher than $40$ dB. It proves that setting different SNRs when generating the synthetic dataset can improve the robustness of the synthetic dataset to noise and make it work better in different SNR environments.

\subsection{Threshold Selection}
In practical applications, the threshold should be adapted to specific requirements. In applications with high security requirements, e.g., financial transaction systems, the primary goal is to prevent attackers from impersonating legitimate users, which requires a higher TPR. According to Fig.~\ref{fig:ROC_synVSexp}, for the proposed CNN-based Siamese model trained on the synthetic dataset, achieving a TPR of 0.95 corresponds to a threshold of 0.64. On the other hand, in user experience-focused applications, e.g., streaming services, the priority is to ensure a smooth experience for legitimate users and reduce the rejection of valid devices, which necessitates a lower FPR. As shown in Fig.~\ref{fig:ROC_synVSexp}, when FPR is set to 0.1, the corresponding threshold is 0.94. 

In summary, selecting an appropriate threshold is crucial for balancing security and usability in authentication systems.

\section{Discussion}
\label{sec:discussion}
\subsection{Scalability Analysis}
The proposed method can be efficiently scaled for large-scale deployment involving multiple devices. In scenarios with multiple attackers ($\text{Mallory}_{1}$, ..., $\text{Mallory}_{n}$) and a legitimate device, as long as the distance between each attacker and the legitimate device exceeds half a wavelength, the packets sent by the attackers and the legitimate device will experience different channel fading. This allows Alice to detect the presence of the attackers based on the estimated CSI measurements. In cases with multiple legitimate users ($\text{Bob}_{1}$, ..., $\text{Bob}_{n}$), Alice can recognize each legitimate user using the same model without the need for retraining.

Although we use the IEEE 802.11 legacy OFDM mode as the focus of our study, the proposed method is applicable to OFDM-based systems that support channel estimation. This makes the method scalable across a wide range of OFDM-based standards, including WiFi, LTE, and 5G.

\subsection{Advantage of Synthetic Dataset}
The proposed synthetic training dataset offers two key advantages as follows.
\begin{itemize}
    \item Efficiency and cost-effectiveness: The synthetic dataset is generated through simulation, eliminating the need for extensive labor-intensive CSI collection. It significantly reduces time and resource requirements, making it more time-efficient and cost-effective.
    \item Enhanced accuracy of channel generation: The synthetic dataset generates more accurate channel conditions than labor-collected datasets, including configurations for SNRs, attack distances, and mobility speeds. It enables the synthetic dataset to thoroughly capture the characteristics of different channel environments, thereby enhancing the generalization capability of the proposed method across different scenarios. As demonstrated in Fig.~\ref{fig:AUC_synVSexp}, across a range of scenarios, the synthetic dataset achieves comparable AUC with experimental datasets collected in matched scenarios.
\end{itemize}

\section{Conclusion}
\label{sec:conclusion}
In this paper, we proposed a novel deep learning-based physical layer CSI authentication for mobile scenarios, enhanced by a synthetic training dataset and CNN-based Siamese network. Specifically, a synthetic training dataset was generated to eliminate the overhead of manually collecting experimental datasets. The autocorrelation and the distance correlation of the channel were first modelled and the dataset was generated based on the WLAN TGn channel model. A CNN-based Siamese network was exploited to learn the temporal and spatial similarity between pairs of CSI estimations and a score can be obtained to measure the difference between the input CSI measurements. Device authentication was achieved by comparing the score to an empirically obtained threshold. 
A unique feature of this paper is a synergistic methodology involving both simulation and experimental evaluation. In particular, the simulation evaluation allowed us to tailor the synthetic dataset parameters.
We explored the effect of SNR, the distance between legitimate and rogue devices and transmission interval on authentication performance. 
We then created a WiFi testbed consisting of an ESP32 kit and two LoPy4 boards and carried out extensive experiments in typical indoor environments.
The experiment results demonstrate the reliability of the synthetic training dataset and the generalization of the proposed scheme. Compared with the FCN-based Siamese network and correlation-based benchmark algorithms, the proposed scheme obtains an average of 0.03 and 0.06 gain of AUC under practical test scenarios, respectively.

\bibliographystyle{IEEEtran}
\bibliography{IEEEabrv,cites}

\begin{IEEEbiography}[{\includegraphics[angle=0,width=1in,clip,keepaspectratio]{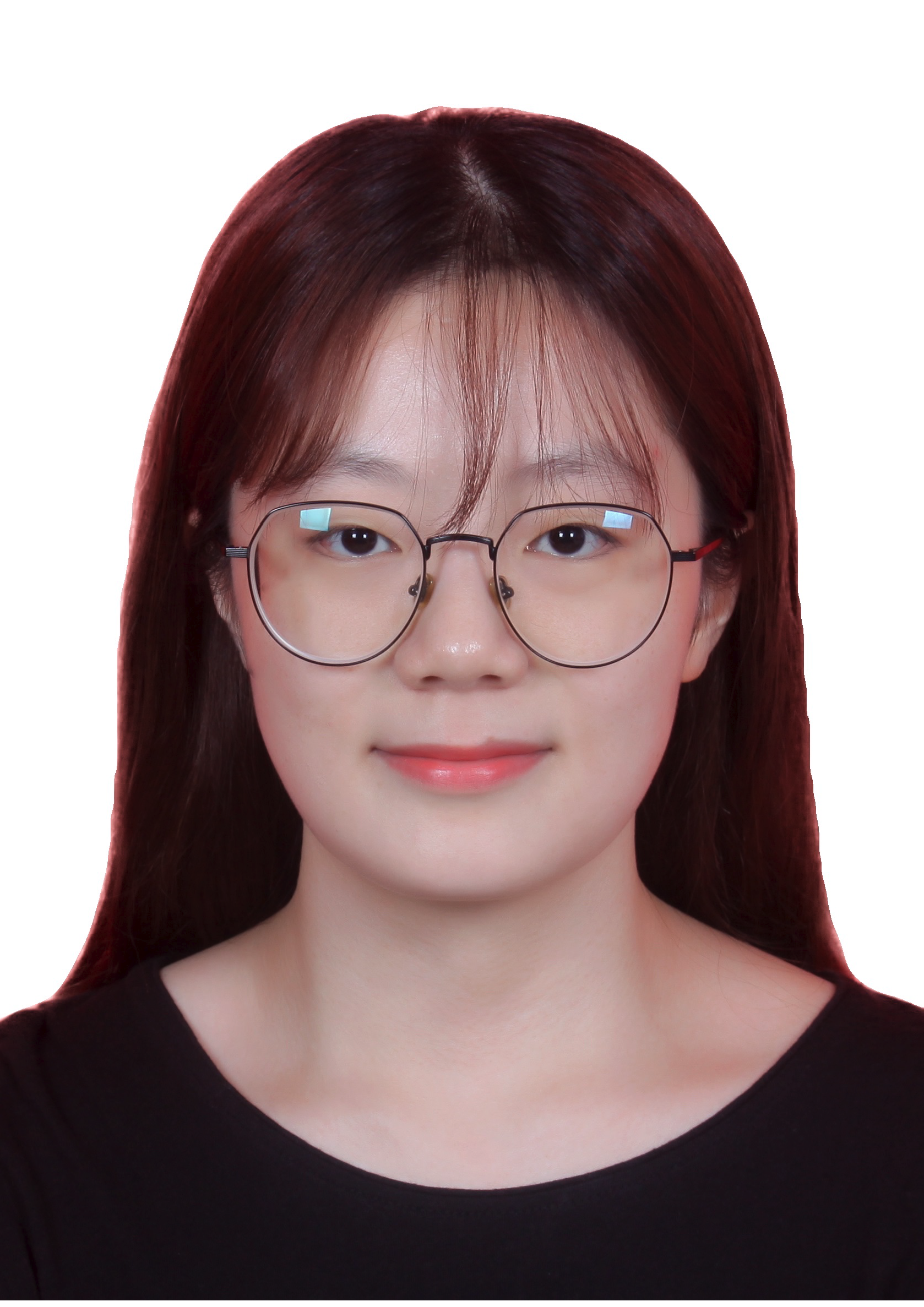}}]{Yijia Guo} (Graduate Student Member, IEEE)
received the M.Eng. degree in information and communication engineering from Sun Yat-sen University, Guangzhou, China, in 2021. She is currently pursuing the Ph.D. degree with the Department of Electrical Engineering and Electronics, University of Liverpool, Liverpool, UK. Her current research interests include the Internet of Things, deep learning and physical layer security.
\end{IEEEbiography}

\begin{IEEEbiography}[{\includegraphics[angle=0,width=1in,clip,keepaspectratio]{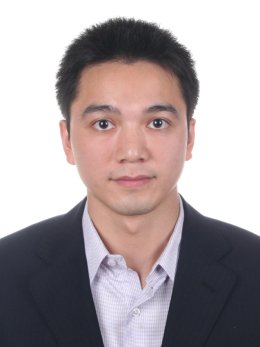}}]{Junqing Zhang}
received B.Eng and M.Eng degrees in Electrical Engineering from Tianjin University, China in 2009 and 2012, respectively, and a Ph.D. degree in Electronics and Electrical Engineering from Queen's University Belfast, UK in 2016. From Feb. 2016 to Jan. 2018, he was a Postdoctoral Research Fellow at Queen's University Belfast. From Feb. 2018 to Oct. 2022, he was a Tenure Track Fellow and then a Lecturer (Assistant Professor) at the University of Liverpool, UK. Since Oct. 2022, he has been a Senior Lecturer (Associate Professor) at the University of Liverpool. His research interests include the Internet of Things, wireless security, physical layer security, key generation, radio frequency fingerprint identification, and wireless sensing. 
Dr. Zhang is a co-recipient of the IEEE WCNC 2025 Best Workshop Paper Award. He is a Senior Area Editor of IEEE Transactions on Information Forensics and Security and an Associate Editor of IEEE Transactions on Mobile Computing.
\end{IEEEbiography}

\begin{IEEEbiography}
[{\includegraphics[angle=0,width=1in,clip,keepaspectratio]{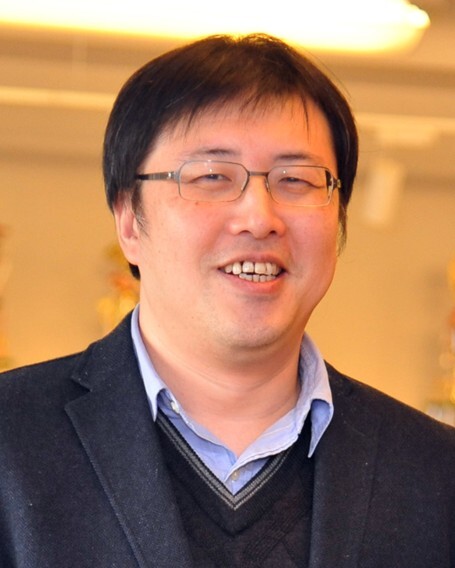}}]{Y.-W. Peter Hong} (S'01 - M'05 - SM'13) 
received the B.S. degree in electrical engineering from National Taiwan University, Taipei, Taiwan, in 1999, and the Ph.D. degree in electrical engineering from Cornell University, Ithaca, NY, USA, in 2005. He is currently a Distinguished Professor with the Institute of Communications Engineering and the Department of Electrical Engineering, National Tsing Hua University (NTHU), Hsinchu, Taiwan. His research interests include AI/ML for wireless communications, UAV and satellite communications, distributed signal processing for the IoT and sensor networks, and physical layer security.

Dr. Hong received the IEEE ComSoc Asia-Pacific Outstanding Young Researcher Award in 2010, the Y. Z. Hsu Scientific Paper Award in 2011, the National Science Council Wu Ta-You Memorial Award in 2011, the Chinese Institute of Electrical Engineering (CIEE) Outstanding Young Electrical Engineer Award in 2012, and the National Science and Technology Council (NSTC) Outstanding Research Award in 2018 and 2022. He was the Chair of the IEEE ComSoc Taipei Chapter from 2017 to 2018. He was the Co-Chair of the Technical Affairs Committee, the Information Services Committee, and the Chapter Coordination Committee of the IEEE ComSoc Asia-Pacific Board, from 2014 to 2015, from 2016 to 2019, and from 2020 to 2021, respectively. He is now the Vice Director of the IEEE ComSoc Asia-Pacific Board from 2022 to 2025. He was also a Distinguished Lecturer of the IEEE Communications Society from 2022 to 2023. In the past, he also served as an Associate Editor for IEEE Transactions on Signal Processing and IEEE Transactions on Information Forensics and Security and an Editor for IEEE Transactions on Communications. He also served as a Senior Area Editor for IEEE Transactions on Signal Processing.
\end{IEEEbiography}

\end{document}